\RequirePackage{fix-cm}
\documentclass[smallextended]{svjour3}       
\smartqed  
\usepackage{url}
\usepackage{graphicx}
\usepackage{array}
\usepackage{multirow}
\usepackage{booktabs}

\usepackage{natbib}
\usepackage{amssymb}
\usepackage{mathtools}
\usepackage{listings,xcolor}
\usepackage{pgffor}

\begin{document}

\title{European Court of Human Rights Open Data project
}


\author{Alexandre Quemy
}


\institute{Alexandre Quemy \at
              IBM, Cracow Software Lab, Poland \\
              Faculty of Computing, Poznan University of Technology, Pozna\'n, Poland\\
              \email{aquemy@pl.ibm.com}           
}

\date{Received: date / Accepted: date}

\maketitle

\begin{abstract}
This paper presents thirteen datasets for binary, multiclass and multilabel classification based on the European Court of Human Rights judgments since its creation. The interest of such datasets is explained through the prism of the researcher, the data scientist, the citizen and the legal practitioner. Contrarily to many datasets, the creation process, from the collection of raw data to the feature transformation, is provided under the form of a collection of fully automated and open-source scripts. It ensures reproducibility and a high level of confidence in the processed data, which is some of the most important issues in data governance nowadays.
A first experimental campaign is performed to study some predictability properties and to establish baseline results on popular machine learning algorithms. The results are consistently good across the binary datasets with an accuracy comprised between 
75.86\% and 98.32\% for an average accuracy of 96.45\%.

\end{abstract}

\section{Introduction}\label{sec:intro}

In this paper, we present the European Court of Human Rights Open Data project (ECHR-OD). It aims at providing up-to-date and complete datasets about the European Court of Human Rights decisions since its creation. To be up-to-date and exhaustive, we developed a fully automated process to regenerate the entire datasets from scratch, starting from the collection of raw documents. As a result, datasets are as complete as they can be in terms of number of cases. The reproducibility makes it easy to add or remove information in future iterations of the datasets. To be able to check for corrupted data or bias, black swans or outliers, the whole datasets generation process is open-source and versioned.

In a second part, we present the results of a large experimental campaign performed on three flavors of the 13 datasets. We compared 13 standard machine learning algorithms for classification with regards to several performance metrics. Those results provide a baseline for future studies and provide some insights about the interest of some types of features to predict justice decisions. Notably, as for previous studies, we found that case textual description contains interesting elements to predict the (binary) outcome. However, for the first time, we show that the judgement is not as good as purely descriptive features to determine what article a given case is about, such that, for real-life predictive systems, the methodology of previous studies might not be suitable by itself.

Before presenting the project and datasets in Section \ref{sec:presentation}, we discuss in Section \ref{sec:context_related_work} the importance of data quality and the multiple issues with current datasets in complex fields such as the legal domain. The creation process is presented in detail in Section \ref{sec:process}. Section \ref{sec:experiments} is dedicated to the experiments on the datasets while Section \ref{sec:conclusion} concludes the papers by discussing the remaining challenges and future work. The paper comes with Supplementary Material available on GitHub\footnote{\url{https://aquemy.github.io/ECHR-OD_project_supplementary_material/}}. It contains additional examples about the data format, as well as all secondary results of the experiments that we omitted due to space constraint.

\section{Context and related work}
\label{sec:context_related_work}

\begin{figure}[!h]
\centering
\includegraphics[scale=0.5]{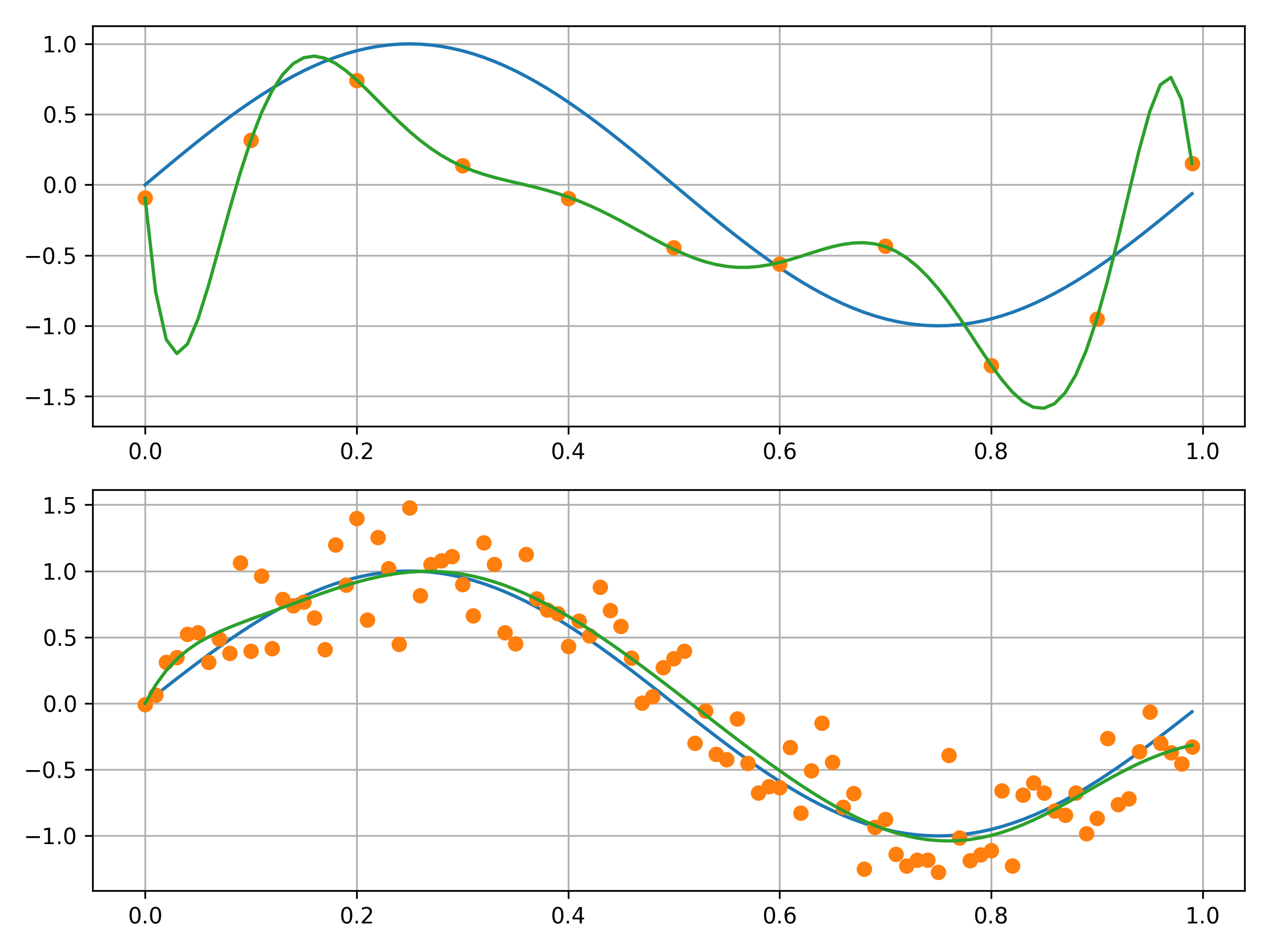}
\caption{{\bf The regularization effect of data}
The ground truth (blue) is a $sin(x)$ modeled by a polynomial of degree 9 (green). On top, with only 11 training points, the model does not approximate correctly the ground truth while at the bottom, with 100 training points, the model error is far lower.}
\label{fig:reg}
\end{figure}

It is now well established that the recent and spectacular results of artificial intelligence, notably with deep learning (\cite{lecun2015deep}), are partly due to the availability of data, so called ``Big Data'', and the exponential growth of computational power. 
For a very long time, the bias-variance tradeoff seemed to be an unbeatable problem: complex models reduce the bias but hurt the variance, while simple models lead to high variance. In parallel, the regularization effect of additional data for complex models was also well known as illustrated by Figure \ref{fig:reg}. The advent of representation learning (\cite{bengio2013representation}) allowed to efficiently build extremely complex models with moderate variance by letting the algorithms discovering the interesting ``patterns'' or ``representations'' to solve a given problem. However, it requires a considerable amount of data to correctly reduce the variance, and there is now a growing consensus on the fact that data are as important as algorithms. In particular, the quality of a model is bounded by the quality of the data it learns from (\cite{valiant1984theory,vapnik1999overview}). The availability and quality of data are thus of primary importance for researchers and practitioners.

Beyond the scope of pure scientific interest, the data governance, that is to say the lifecycle management of the data, is particularly crucial for our modern societies (\cite{OLHEDE201846,tallon2013corporate,attard2018challenges}). What data are publicly available? Who produces, manages and manipulates those data? What is the quality of the data? What is the process of collection, curation and transformation? Those are few legitimate questions that a citizen, a company, or an institution may (should?) ask due to the ethical, political, social and legal concerns (\cite{kitchin2014data}). One can mention the recent General Data Protection Regulation\footnote{\url{https://www.eugdpr.org/}} (GDPR), a European Union law with global application, that tries to give a legal framework to address some of the abovementioned questions. Beside privacy and business considerations, the quality of data is at the core of the quality of insights and decisions derived from the data.

\subsection{Open Data limits and validity threat}

The Open Data movement considers that the data should be freely available and reusable by anyone (\cite{kitchin2014data}). Although it has strong beneficial effects at many levels (for instance, in science (\cite{10.1371/journal.pbio.1001195}), in civic engagement (\cite{kassen2013promising}), or in governmental transparency (\cite{janssen2012benefits}), some critics have emerged with regards to social questions (\cite{gurstein2011open,misuraca2014open}). Aside from those specific considerations, we argue that open data are not enough to insure data quality and to totally handle all the questions mentioned in the previous paragraph. There are many ways data may be unsuitable for making decisions (either solely human based or assisted by any kind of model):
\begin{itemize}
\item {\bf Data sparsity and irrelevant information:} a dataset may lack information to correctly make a decision or on the contrary, contain a lot of irrelevant information. Some useful information might not be available at the moment the dataset is constructed. Also, it is hard to know {\it a priori} what piece of information is useful or not to model or understand a phenomenon. It usually requires specific techniques such as feature selection (\cite{guyon2003introduction}) and several studies to obtain a big picture. Having a dataset, even open, without the whole process from the collection to the moment it is publicly available is usually not enough for practical applications.
\item {\bf Missing unexpected patterns or learning wrong patterns:} for some reasons, regime change might occur in the data and be learnt or not by a model. How to know if this change is valid or is the result of a problem somewhere between the data collection (e.g. some sensors are not working or being recalibrated) and the data processing (e.g. a bug in the software used to sanitize the data) {\bf without} expert knowledge? Some points may also be outliers for good reasons (e.g. improbable event that {\it eventually} occurs, often referred as a {\it black swan}) or bad reasons (e.g. error in processing the data, problem in collecting the data)? In the first case, the models {\bf must} take into account those data, while in the second case, it should be discarded. Open data cannot help, except for obvious cases.
\item {\bf Data corruption:} at any stage of the collection, processing and usage, the data may be partly corrupted. It may be hard to determine where and how the data has been corrupted even if the data are open.
\item {\bf Biased data:} from the collection process itself to the sanitization choices, bias is introduced. Having access to open data is no help for building better models and algorithms if data are biased from the beginning.
\end{itemize} For those reasons, the datasets presented in this paper are accompanied with the full creation process, carefully documented. 

Let us illustrate some of those limitations through a concrete case that motivated this project. To validate and compare a new method for classification, we used some datasets provided along with an article published in an open data journal. The reproduction of the experiment was successful but the results with the new methods were inconsistent with our preliminary tests. By digging into the datasets, we found out that many input vectors were empty (up to 70\% for some datasets) and the labels were not consistent among those empty vectors. In other words, many situations to classify were described with no information, and this absence of information could not be linked to a specific label. Furthermore, the prevalence in the group of degenerated vectors was rather high in comparison to the overall prevalence, resulting in relatively good metrics for standard classification algorithms. As the authors did not provide the raw data and the transformation process, there was no way to figure out what exactly caused this problem.

We would like to insist on the last two elements, namely the data corruption and biased data, as they represent a huge validity threat in machine learning-based retrospective studies. The concern is articulated around the following: 
\begin{enumerate}
\item a typical study focuses on a dataset transformed by a data pipeline. For a new method, a cross-validation is used with the preprocessed data. Results are used to make a comparison with other methods presented in previous studies.
\item in general, comparisons are done solely between methods, without taking into account the data pipeline.
\item however, the data preprocessing operations introduces bias and possibly data corruption, which can drastically affect the final results.
\end{enumerate}
It may be difficult to evaluate how much the effect of data preprocessing affects the final result, especially that the data pipelines are rarely reported. For instance, \cite{CRONE2006781} notice that algorithm hyperparameter tuning is performed in 16 out of 19 selected publications while only two publications study the impact of data preprocessing.

The data preprocessing impact has been evaluated for multiple algorithms and operators. In \cite{CRONE2006781}, the authors showed that the accuracy obtained by Neural Network, SVM and Decision Trees are significantly impacted by data scaling, sampling and continuous and categorical coding. A correlation link between under and oversampling is also demonstrated. In \cite{NAWI201332}, three specific data processing operators has been tested for neural networks. Despite the authors do not provide the results without any data processing, the results show an important accuracy variability between the alternatives, thus implying a data processing impact. Using a representative sample of the available data can also lead to better overall results, as showed by \cite{DBLP:conf/bdas/NalepaMPHK18}. For a more comprehensive view on data processing impact, we refer the reader to \cite{dasu2003exploratory}. Recently, \cite{DBLP:conf/dolap/Quemy19} focuses on the data pipeline optimization and found that, between no preprocessing step and a carefully selected data pipeline, the classification error is reduced by 66\% in average among four methods (SVM, Decision Tree, Neural Network and Random Forest) and three datasets (Iris, Breast, Wine). More interesting, by changing the data pipeline, it is possible to obtain any possible ranking of the methods with regard to the error rate.

\subsection{Related work in legal analytics}

The legal environment is a {\it messy concept} (\citet{1637349}) that intrinsically poses some of the most challenging problems for the artificial intelligence research community: grey areas of interpretation, many exceptions, non-stationarity, presence of deductive and inductive reasoning, non-classical logic, etc. For some years and in several areas of the law, some "quantitative" approaches have been developed, based on the use of more or less explicit mathematical models. With the availability of massive data, those trends have been accented and brand-new opportunities are emerging at a sustained pace. Among the stakes of those studies, one can mention a better understanding of the legal system and the consequences of some decisions on the society, but also the possibility to decrease the mass of litigations in a context of cost rationalization. For a survey on legal analytics methods, we refer the reader to \citet{DBLP:conf/adbis/Quemy17}.
 In \citet{DBLP:conf/adbis/Quemy17}, the author also defines some practical problems in the field of legal analytics. First, the prediction problem consists in determining the outcome of a trial given some facts about the specific case and some knowledge about the legal environment. The second problem consists in building a legal justification, knowing the legal environment.

The justification problem should not be misunderstood with the model explainability problem in machine learning. Understanding how a model make a prediction is certainly useful to generate legal justifications, however it is not enough. Many studies modeled justice decisions solely based on the estimation of the judge ideology (\citet{33cbff13d8b040ec8827ac230aed9caf,segal1989,doi:10.2307/2960194,citeulike:1147857}). The way the model makes decisions is rather clear, however, the model is unable to provide a satisfying legal justification. Those considerations are highly connected to a major debate among the legal practitioners, namely legalism to realism: are the judges objectively applying a method contained in the text (legalism) or do they create their own interpretation (realism). The feasibility of a solution to the justification problem largely depends on the answer to this debate. Some discussions on the topic can be found in \citet{posner2010judges,JELS:JELS1255,tamanaha2012balanced,leiter2010legal}.

By releasing large datasets, with several flavors based on different types of features, along with the whole tunable preprocessing pipeline, we hope to gain a better understanding on how justice decisions are taken, what elements are useful for a prediction, and to quantify the balance between realism and legalism. We also hope to make a step toward solving the justification problem.

Predicting the outcome of a justice case is challenging, even for the best legal experts: 67.4\% and 58\% accuracy, respectively for the judges and whole case decision, is found in \cite{PCS3} for the Supreme Court of the United States (SCOTUS). Using crowds, the Fantasy Scotus\footnote{\url{https://fantasyscotus.lexpredict.com/}} project reached respectively 85.20\% and 84.85\% correct predictions.

 The Supreme Court of the United States (SCOTUS) has been widely studied, notably through the SCOTUS database\footnote{\url{http://scdb.wustl.edu/}} (\citet{Katz2017,martin_quinn_ruger_kim_2004,Guimer2011}). This database is composed of structured information about every case since the creation of the court but no textual information from the opinions. The opinions and other related textual documents also have been studied separately for SCOTUS (\citet{Islam,AJPS,Sim}). Conversely, very little if no similar work has been done in Europe. As far as we know, the only predictive model was using the textual information only (\citet{10.7717/peerj-cs.93}), despite more structured information is publicly available on HUDOC\footnote{\url{https://hudoc.echr.coe.int/eng}}. Using NLP techniques, the authors of \cite{10.7717/peerj-cs.93} achieve 79\% accuracy to predict the decisions of the European Court of Human Rights (ECHR). They make the hypothesis that the textual content of the European Convention of the Human Rights holds hints that will influence the decision of the Judge. They extracted from the judgement documents the top 2000 N-grams, calculated their similarity matrix based on cosine measure and partition this matrix using Spectral Clustering to obtain a set of interpretable topics. The binary prediction was made using an SVM with linear kernel. Contrary to the previous studies on SCOTUS, they found out that the formal facts are the most important predictive factors, which tend to favor legalism.

 The data used in \citet{10.7717/peerj-cs.93} are far from being exhaustive: 3 articles considered (3, 6 and 8) with respectively 250, 80 and 254 cases per article. For many data-driven algorithms, this might be too little to build a correct model. As far as we know, the project presented in this paper already provides the largest existing legal datasets directly consumable by machine learning algorithms. In particular, it includes several types of features: purely descriptive and textual.

\subsection{On the availability of legal data}

One of the prominent domains of application for deep learning is computer vision. In this area, it is relatively easy to obtain new data, even if the process of manually labelling training data may be laborious. Techniques like Data Augmentation allow to generate artificially new data by slightly modifying existing examples (\citet{doi:10.1198/10618600152418584}). For instance, for hand-written recognition, one may add some small perturbations or apply transformations to an existing example, such as rotation, zoom, gaussian noise, etc. Those techniques are efficient at providing useful new examples but rely on an implicit assumption: the solution's behavior changes continuously with the initial data. In other words, slightly modifying the picture of an 8 by a small rotation or distortion still results in an 8. However, in many other fields, a small change in the data may result in a totally different outcome such that one cannot use Data Augmentation to artificially grow her dataset. In general, the fields where Data Augmentation is not applicable are more complex, require more information to process or require a sophisticated or {\it conscious} reasoning before being able to give an answer. Anyone can recognize a cat from a horse without processing any additional data than the picture itself, without elaborating a complex and explicit reasoning. Conversely, deciding if someone is guilty w.r.t. some available information and the current legal environment is not as natural as recognizing a cat, even for the best legal experts.

We would like to draw the attention on the fact that, it is not because those fields are more complex for humans that nowadays artificial intelligence techniques cannot perform better than humans: medical diagnosis is a complex field requiring expertise and explicit reasoning, however humans are regularly beaten by the machine (\citet{Tiwari2016,Yu2016,Patel2016}). That said, in many complex fields, data augmentation techniques cannot be used, access to the data may be difficult, the data itself can be limited and, when open, the data are provided already curated without access to the curation process. ECHR-OD project has been created with those considerations in mind and we will see with the experiments that despite being exhaustive w.r.t. number of cases, more data would have helped the models to perform better.

\section{ECHR-OD in brief}
\label{sec:presentation}

{\bf ECHR-OD} project aims at providing exhaustive and high-quality database and datasets for diverse problems, based on the European Court of Human Rights documents available on HUDOC. It appears important to us 1) to draw the attention of researchers on this domain that has important consequences on the society, 2) to provide a similar and more complete database for the European Union as it already exists in the United States, notably because the law systems are different in both sides of the Atlantic.
The project is composed of five components:
\begin{enumerate}
\item {\bf Main website:} \url{https://echr-opendata.eu}
\item {\bf Download mirror:} \url{https://osf.io/52rhg/}
\item {\bf Creation process:} \url{https://github.com/aquemy/ECHR-OD_process}
\item {\bf Website sources:} \url{https://github.com/aquemy/ECHR-OD_website}
\item {\bf Data loader in python:} \url{https://github.com/aquemy/ECHR-OD_loader}
\end{enumerate}

ECHR-OD is guided by three core values: {\bf reusability}, {\bf quality} and {\bf availability}. To reach those objectives,
\begin{itemize}
\item each version of the datasets is carefully versioned and publicly available, including the intermediate files,
\item the integrality of the process and files produced are careful documented,
\item the scripts to retrieve the raw documents and build the datasets from scratch are open-source and carefully versioned to maximize reproducibility and trust,
\item no data is manipulated by hand at any stage of the creation process.
\end{itemize}

At the submission date, the project offers 13 datasets for the classification problem. Datasets for other problems such as structured predictions will be available in the future. The datasets are available under the {\bf Open Database Licence (ODbL)}\footnote{Summary: \url{https://opendatacommons.org/licenses/odbl/summary/}}\footnote{Full-text: \url{https://opendatacommons.org/licenses/odbl/1.0/}} which guarantees the rights to copy, distribute and use the database, to produce work from the database and to modify, transform and build upon the database. The creation scripts and website sources are provided under {\bf MIT Licence}.

\subsection{Datasets description}

In machine learning, the problem of classification consists in finding a mapping from an input vector space $\mathcal X$ to a discrete decision space $\mathcal Y$ using a set of examples. The binary classification problem is a special case of the multiclass such that $\mathcal Y$ has only two elements, while in multilabel classification, each element of $\mathcal X$ can have several labels. It is often viewed as an approximation problem s.t. we want to find an estimator $\bar J$ of an unknown mapping $J$ available only through a sample called {\it training set}. A training set $(\mathbf{X}, \mathbf{y})$ consists of $N$ input vectors $\mathbf{X} = \{ \mathbf{x}_1, ..., \mathbf{x}_N \}$ and their associated correct class $\mathbf{y} = \{ J(\mathbf{x}_i) + \varepsilon \}^{N}_{i=1}$, possibly distorted by some noise $\varepsilon$. Let $\mathcal{J}(\mathcal X, \mathcal Y)$ be the class of mappings from $\mathcal X$ to $ \mathcal Y$. Solving an instance of the classification problem consists in minimizing the classification error:

\begin{equation}
J^* = \underset{\bar J \in \mathcal{J}(\mathcal X, \mathcal Y)}{arg\,min} \sum_{\mathbf{x} \in \mathcal X} \mathbb{I}_{\{J(\mathbf x) \neq \bar J(\mathbf{x})\}}
\end{equation}

From the HUDOC database and judgment files, we created several datasets for three variants of the classification problem: binary classification, multiclass classification and multilabel classification. There are 11 datasets for binary, one for multiclass, and one for multilabel classification.

Each dataset comes in different flavors based on descriptive features and Bag-of-Words and TF-IDF representations:
\begin{enumerate}
  \item {\bf Descriptive features:} structured features retrieved from HUDOC or deduced from the judgment document,
  \item {\bf Bag-of-Words representation:} based on the top 5000 tokens (normalized $n$-grams for $n\in \{1,2,3,4\}$),
  \item {\bf TF-IDF representation:} idem but with a TF-IDF transformation to weight the tokens,
  \item {\bf Descriptive features + Bag-of-Words:} combination of both sets of features,
  \item {\bf Descriptive features + TF-IDF:}  combination of both sets of features.
\end{enumerate}
Those different representations exist to study the respective importance of descriptive and textual features in the predictive models build upon the datasets.

For binary classification, the label corresponds to a violation or no violation of a specific article. Each of the 11 datasets corresponds to a specific article. We kept only the articles such that there are at least 100 cases with a clear output (see Section 
\ref{sec:filter} for additional details) without consideration on the prevalence. Notice that a same case can appear in two datasets if it has in his conclusion two elements about a different article. A basic description of those datasets is given by Table \ref{table:binary}.

\begin{table}
  \caption{{\bf Datasets description for binary classification.}}
  \label{table:binary}
  \begin{tabular}{@{} lrrrrr@{} }
\toprule
 & \# cases & min \#features & max \#features & avg \#features & prevalence \\ \midrule
Article 1 & 951 & 131 & 2834 & 1183.47 & 0.93\\
Article 2 & 1124 & 44 & 3501 & 2103.45 & 0.90\\
Article 3 & 2573 & 160 & 3871 & 1490.75 & 0.89\\
Article 5 & 2292 & 200 & 3656 & 1479.60 & 0.91\\
Article 6 & 6891 & 46 & 3168 & 1117.66 & 0.89\\
Article 8 & 1289 & 179 & 3685 & 1466.52 & 0.73\\
Article 10 & 560 & 49 & 3440 & 1657.22 & 0.75\\
Article 11 & 213 & 293 & 3758 & 1607.96 & 0.85\\
Article 13 & 1090 & 44 & 2908 & 1309.33 & 0.91\\
Article 34 & 136 & 490 & 3168 & 1726.78 & 0.64\\
Article p1 & 1301 & 266 & 2692 & 1187.96 & 0.86\\
\bottomrule
\end{tabular}
  \begin{flushleft}  The columns  min, max, avf \#features indicate the minimal, maximal and average number of features in the dataset cases for the representation "descriptive features + Bag-of-Words representation".
  \end{flushleft}
\end{table}

For multiclass, there is a total of 18 different classes (the number of different articles multiplied by two possible decisions: violation or no violation).
To create the multiclass dataset, we aggregate the different binary classification datasets by removing the cases present in several datasets. For this reason, articles 13 and 34 are not included since they had less than 100 cases after this step. We did not simply merge the binary classification datasets for the remaining cases. The processing part consisting in the creation of the BoW and TF-IDF representations is based on the 5000 most frequent $n$-grams among the corpus of judgments (see Section \ref{sec:normalization} documents for additional details). As the most frequent $n$-gram changes depending on the corpus, the BoW representation of a given case is different in the binary, multiclass and multilabel dataset. The descriptive features are, however, not modified. 

\begin{figure}[!h]
\centering
\includegraphics[scale=0.4]{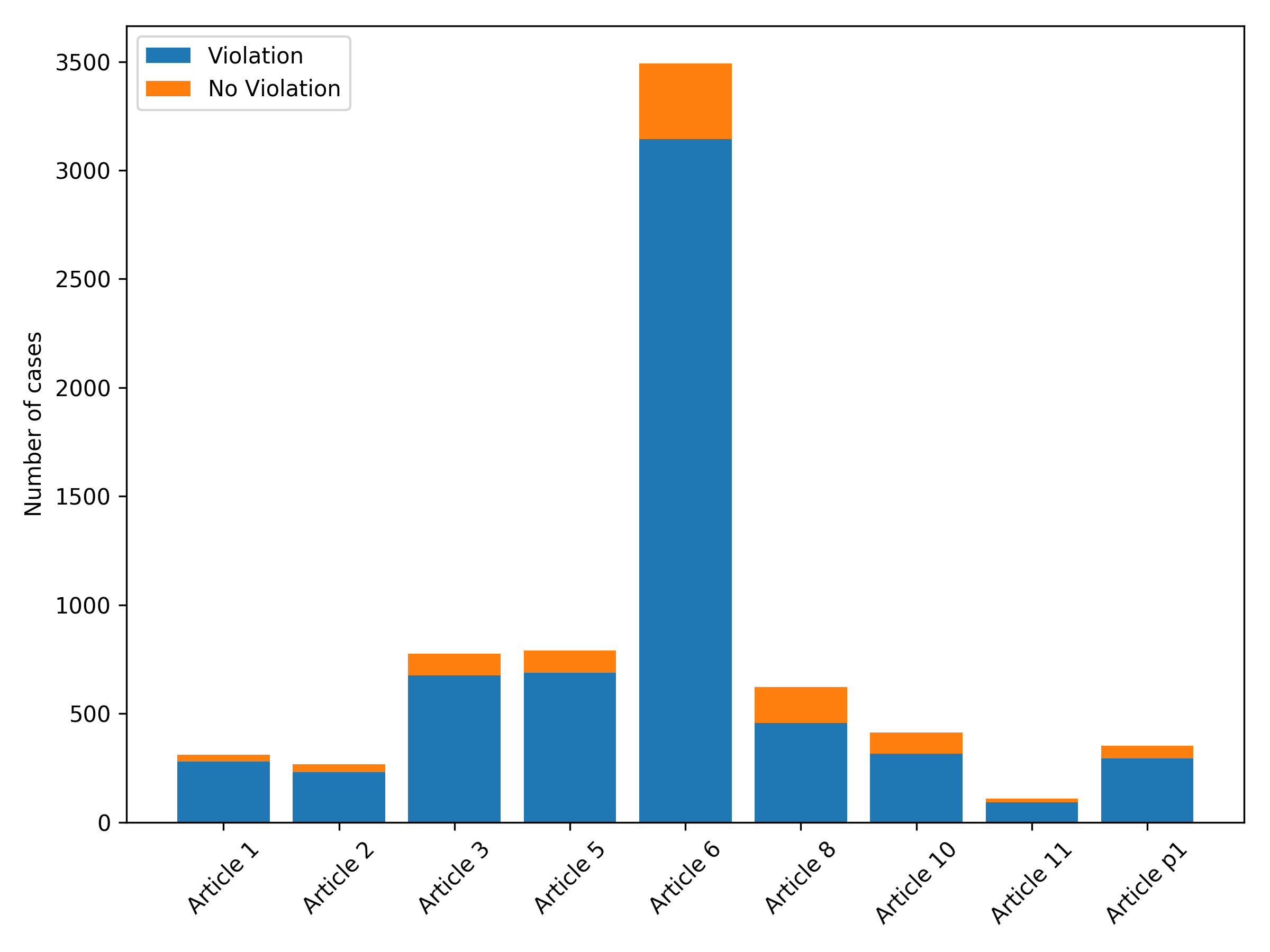}
\caption{{\bf Number of cases depending on the article and the outcome for the multiclass dataset.}}
\label{fig:multiclass_count}
\end{figure}

\begin{figure}[!h]
\centering
\includegraphics[scale=0.4]{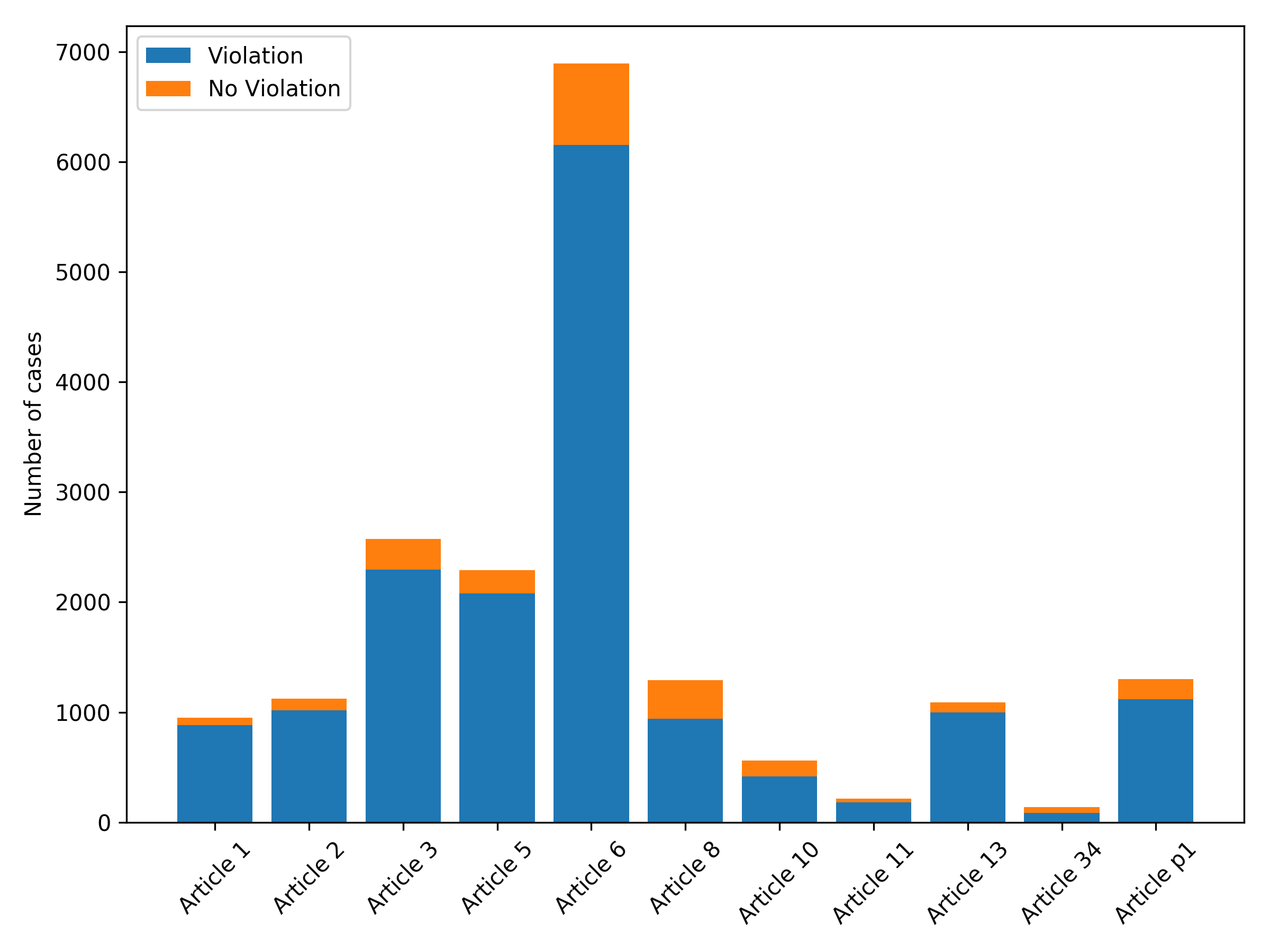}
\caption{{\bf Number of cases depending on the article and the outcome for the multilabel dataset.}}
\label{fig:multilabel_count}
\end{figure}

\begin{figure}[!h]
\centering
\includegraphics[scale=0.4]{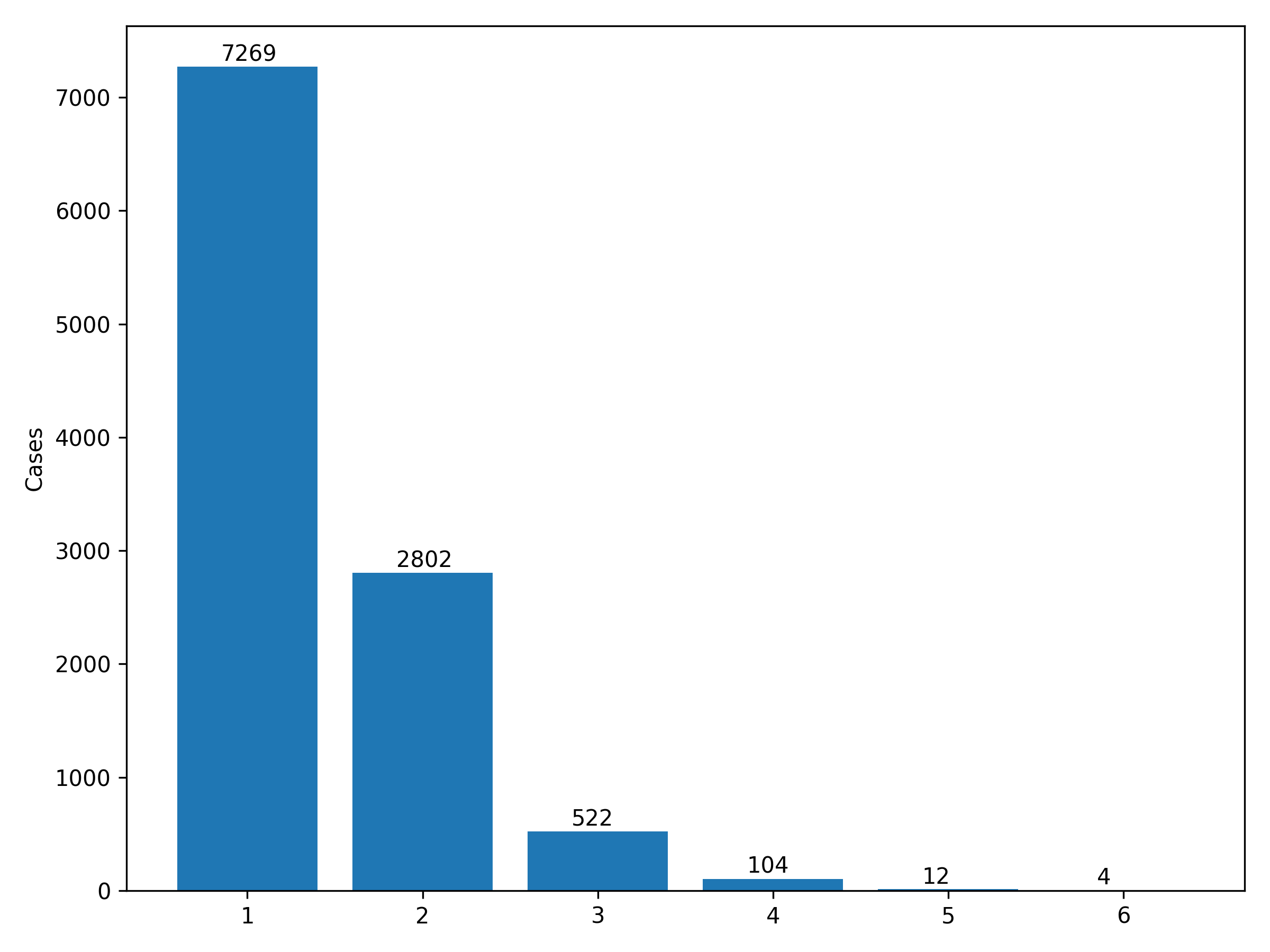}
\caption{{\bf Number of cases depending on the number of labels for the multilabel dataset.}}
\label{fig:multilabel_count_labels}
\end{figure}

For multilabel classification, there are 22 different labels and the main difference with the multiclass is that there is no need to remove cases that appear in multiple binary classification datasets. The labels are simply stacked. Table \ref{table:multilabel} summarizes the dataset composition, Figures  \ref{fig:multiclass_count} and \ref{fig:multilabel_count} shows the labels repartition among the multiclass and multilabel datasets, and Figure \ref{fig:multilabel_count_labels} provides the histogram of label numbers and cases per label. 

\begin{table}
  \caption{{\bf Dataset description for the multiclass dataset.}}
  \label{table:multiclass}
  \centering
  \begin{tabular}{@{} lrrrrr@{} }
\toprule
 & \# cases & violation & no-violation & prevalence\\ \midrule
Article 1 & 310 & 280 (0.039) & 30 (0.004) & 0.90 \\
Article 2 & 267 & 230 (0.032) & 37 (0.005) & 0.86 \\
Article 3 & 775 & 676 (0.095) & 99 (0.014) & 0.87 \\
Article 5 & 791 & 689 (0.097) & 102 (0.014) & 0.87 \\
Article 6 & 3491 & 3143 (0.441) & 348 (0.049) & 0.90 \\
Article 8 & 623 & 457 (0.064) & 166 (0.023) & 0.73 \\
Article 10 & 413 & 315 (0.044) & 98 (0.014) & 0.76 \\
Article 11 & 110 & 92 (0.013) & 18 (0.003) & 0.84 \\
Article p1 & 353 & 294 (0.041) & 59 (0.008) & 0.83 \\
\bottomrule
\end{tabular}
  \begin{flushleft}For each article is indicated the number of cases, the number of cases labeled as violated and not violated with in parenthesis the prevalence w.r.t. the {\bf whole} dataset.
  \end{flushleft}
\end{table}

\begin{table}
  \caption{{\bf Dataset description for the multilabel dataset.}}
  \label{table:multilabel}
  \centering
  \begin{tabular}{@{} lrrrrr@{} }
\toprule
 & \# cases & violation & no-violation \\ \midrule
Article 1 & 951 & 882 (0.082) & 69 (0.006) \\
Article 2 & 1124 & 1017 (0.095) & 107 (0.010) \\
Article 3 & 2573 & 2295 (0.214) & 278 (0.026) \\
Article 5 & 2292 & 2081 (0.194) & 211 (0.020) \\
Article 6 & 6891 & 6152 (0.574) & 739 (0.069) \\
Article 8 & 1289 & 940 (0.088) & 349 (0.033) \\
Article 10 & 560 & 418 (0.039) & 142 (0.013) \\
Article 11 & 213 & 180 (0.017) & 33 (0.003) \\
Article 13 & 1090 & 997 (0.093) & 93 (0.009) \\
Article 34 & 136 & 87 (0.008) & 49 (0.005) \\
Article p1 & 1301 & 1120 (0.105) & 181 (0.017) \\
\bottomrule
\end{tabular}
  \begin{flushleft}For each article is indicated the number of cases, the number of cases labeled as violated and not violated with in parenthesis the prevalence w.r.t. the {\bf whole} dataset.
  \end{flushleft}
\end{table}

\begin{table}[!h]
\caption{{\bf Files contained in a dataset.}}
\label{table:dataset_content}
\begin{center}
\begin{tabular}{@{}ll@{}}
\toprule
{\bf descriptive.txt}  & Descriptive features only. \\
{\bf BoW.txt}  & Bag-of-Word representation only. \\
{\bf TF\_IDF.txt}  & TF-IDF representation only.  \\
{\bf descriptive+BoW.txt} & Descriptive features and Bag-of-Words.  \\
{\bf descriptive+TF\_IDF.txt}  & Descriptive features and TF-IDF.  \\
{\bf outcomes.txt}  & Contain the labels of the datasets. \\
{\bf features\_descriptive.json} & Mapping between feature and numerical id.\\
{\bf features\_text.json} & Mapping between $n$-grams and numerical id.\\
{\bf outcomes\_variables.json} & Mapping between labels and numerical id. \\
{\bf variables\_descriptive.json} & Mapping between descriptive variable and numerical id.\\
{\bf statistics\_datasets.json} & Contain some statistics about the dataset.   \\ \bottomrule
\end{tabular}
\end{center}
\end{table}

The final format to encode the case information is close to the LIBSVM format. Each couple (\texttt{variable}, \texttt{value}) is encoded by \texttt{<variable\_id>:<value\_id>} with the specificity that the \texttt{<value\_id>} is not encoded per variable but globally. For instance, \texttt{0:7201} corresponds to variable \texttt{itemid=001-170361}. 
The encoding for the variables can be found in {\bf variables\_descriptive.json} and the encoding for the couples (\texttt{variable}, \texttt{value}) in {\bf features\_descriptive.json}. The format for the mapping {\bf features\_descriptive.json} is \texttt{"<variable>=<value>":<id>} or \texttt{"<variable>\_has\_<value>":<id>} if the variable is a set of elements. For instance, the variable \texttt{parties} has two elements and is encoded by 19. Having "BASYUK" in the parties of a case is encoded by \texttt{"parties\_has\_BASYUK": 109712} and thus, the case description contains \texttt{19:109712} and \texttt{19:X} where X is the id for the second party.
As the id is global, having the variable id in prefix is redundant. Notice that it has at least three advantages. First, there is no need to look in the global dictionary and parse the corresponding key to know the encoded variable. Second, some algorithms might want a pair (\texttt{variable}, \texttt{value}) (e.g. Decision Tree) while others can work with global tokens (e.g. Neural Network). Finally, it makes it easier to re-encode the cases with a specific encoder (e.g. binary, Helmert, Backward Difference, etc.).

Regarding the Bag-of-Words representation, each $n$-gram is turned into a variable such that when a case judgment contains a specific token, the final representation contains \texttt{<token\_id>:<occurrences>}. For instance, assuming that the $2$-gram "find\_guilty" is encoded by 128210 and appears five times in a judgment, the case description will contain \texttt{128210:5}. For TF-IDF representation, \texttt{<occurrences>} is replaced by the specific weight for this token in the document given the whole dataset.

A Python library to load and manipulate the datasets have been developed and is available at \url{https://github.com/aquemy/ECHR-OD_loader}.

\section{Creation process}
\label{sec:process}

\begin{figure}[!h]
\centering
\includegraphics[scale=0.5]{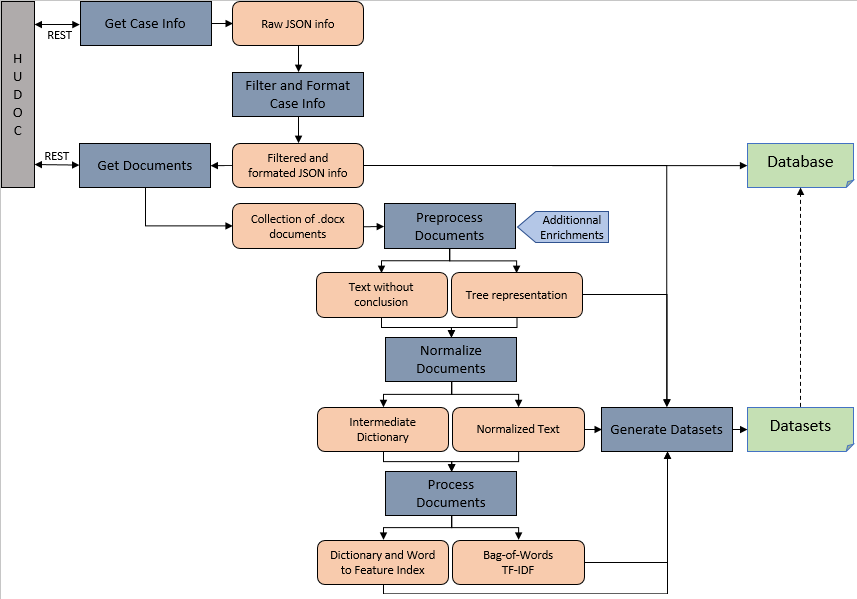}
\caption{{\bf ECHR-OD datasets creation process.}}
\label{fig:process_graph}
\end{figure}

In this section, we describe in detail the dataset generation process from scratch. The datasets are based on the raw documents and information available publicly in HUDOC database. The process is broken down into several steps as illustrated by Figure \ref{fig:process_graph}:

\begin{enumerate}
\item {\bf get\_cases\_info.py:} Retrieve the list and basic information about cases from HUDOC,
\item {\bf filter\_cases.py:} Remove unwanted, inconsistent, ambiguous or difficult-to-process cases,
\item {\bf get\_documents.py:} Download the judgment documents for the filtered list of cases,
\item {\bf preprocess\_documents.py:} Analyze the raw judgments to construct a JSON nested structures representing the paragraphs,
\item {\bf process\_documents.py:} Normalize the documents and generate a Bag-of-Words and TF-IDF representation,
\item {\bf generate\_datasets.py:} Combine all the information to generate several datasets.
\end{enumerate}

The integrality of this process is wrapped into a script {\bf build.py}. This script has some parameters such as the output folder name but also the number of tokens to take into consideration during the generation of Bag-of-Words representation. This allows anyone to generate slightly modified versions of the datasets and to experiment with them. 

\subsection{Retrieving cases}

Using HUDOC API, basic information about all entries are retrieved and saved in JSON files. Those entries contain several keys that are listed on top of the script {\bf get\_case\_info.py}. Among them can be found the case name, the language used, or the conclusion in natural language.
See Appendix \ref{S1_Appendix} for an example of a case description.

\subsection{Filtering cases}
\label{sec:filter}

To ensure the quality and usability of the datasets, we filter the cases as follows:

\begin{enumerate}
  \item We keep only cases in English,
  \item We keep only cases with a judgment document,
  \item We remove the cases without an attached judgment document,
  \item We keep only the cases with a clear conclusion (i.e. containing at least one occurrence of ``(no) violation''),
  \item We remove a specific list of cases hard to process (three cases for this version of the datasets).
\end{enumerate}

During this step, we also parse and format some raw information: the parties are extracted from the case title and many raw strings are broken down into lists. In particular, the string listing the articles discussed in a case are turned into a list and the conclusion string into a slightly more complex JSON object. For instance, the string \texttt{Violation of Art. 6-1;No violation of P1-1;Pecuniary damage - claim dismissed;Non-pecuniary damage - financial award} becomes the list of elements described in Appendix \ref{S2_Appendix}.

In general, each item in the conclusion can have the following elements:
\begin{enumerate}
 \item {\bf article:} number of the concerned article if applicable,
 \item {\bf details:} list of additional information (paragraph or aspect of the article),
 \item {\bf element:} part of the raw string describing the item,
 \item {\bf mentions:} diverse mentions (quantifier s.a. 'moderate', country...),
 \item {\bf type:} violation, no violation or other.
\end{enumerate}
Some representative examples are provided in Appendix \ref{S3_Appendix}.

Finally, on top of saving the case information in a JSON file, we output a JSON file for each unique article with at least 100 associated cases\footnote{This constant is a parameter of the script and can thus be modified for additional experimentations.}.

Additionally, some basic statistics about the attributes are generated, e.g. the cardinality of the domain and the density (i.e. the cardinality over the total number of cases). For instance, the attribute \texttt{itemid} is unique and thus, as expected, its density is 1:

{\small
\begin{lstlisting}
"itemid": {
  "cardinal": 12075, 
  "density": 1.0
}
\end{lstlisting}
}

In comparison, the field \texttt{article\_} (raw string containing a list of articles discussed in a case) and \texttt{article} (its parsed and formatted counterpart) have a density of respectively 25\% and 1\%. This illustrates the interest of our processing method: using the raw string, the article attribute is far more unique than it should be. In reality, there are about 130 different values that are really used across the datasets.
{\small
\begin{lstlisting}
"articles_": {
  "cardinal": 3104, 
  "density": 0.2570600414078675
}
\end{lstlisting}
}

{\small
\begin{lstlisting}
"article": {
  "cardinal": 131, 
  "density": 0.010848861283643893
}
\end{lstlisting}
}

\subsection{Getting documents}

During this phase, we only download the judgment documents in Microsoft Word format using HUDOC API.

\subsection{Preprocessing documents}

The preprocessing step consists in parsing the MS Word document to extract additional information and create a tree structure of the judgment file. It outputs two files for each case:
\begin{enumerate}
\item {\bf \textless case\_id\textgreater\_parsed.json:} same JSON document as produced by {\bf filter\_cases.py} with additional information.
\item {\bf \textless case\_id\textgreater\_text\_without\_conclusion.txt:} full judgment text without the conclusion. It is meant to be used for creating the BoW and TF-IDF representations.
\end{enumerate}

To the previous information, we add the field \texttt{decision\_body} with the list of persons involved into the decision, including their role. See Appendix \ref{S4_Appendix} for an example.

The most important addition to the case info is the tree representation of the whole judgment document under the field \texttt{content}. The content is described in an ordered list where each element has two fields: 1) \texttt{content} to describe the element (paragraph text or title) and 2) \texttt{elements} that represents a list of sub-elements. For a better understanding, see the example in Appendix \ref{S5_Appendix}. This representation eases the identification of some specific sections or paragraphs.

\subsection{Normalizing documents}
\label{sec:normalization}

During this step, the documents {\bf \textless case\_id\textgreater\_text\_without\_conclusion.txt} are normalized as follows:

\begin{itemize}
\item Tokenization,
\item Stopwords removal,
\item Part-of-Speech tagging followed by a lemmatization,
\item $n$-gram generation for $n \in \{1,2,3,4\}$,
\end{itemize}
The output files are named {\bf \textless case\_id\textgreater\_normalized.txt}.

\subsection{Processing documents}
\label{sec:processing}
This step uses Gensim (\citet{rehurek_lrec}) to construct a dictionary of the 5000 most common tokens based on the normalized documents (the dictionary is created per dataset) and outputs the Bag-of-Words and TF-IDF representations for each document. The naming convention is {\bf \textless case\_id\textgreater\_bow.txt} and {\bf \textless case\_id\textgreater\_tfidf.txt}. Additionally, {\bf feature\_to\_id.dict} and {\bf dictionary.dict} contain the mapping between tokens and id, respectively in JSON and in a compressed format used by Gensim. The number of tokens to use in the dictionary is a parameter of the script.

\subsection{Generating datasets}

The final step consists in producing the dataset and related files. See Table \ref{table:dataset_content} for the list of output files.
The feature id of the BoW and TF-IDF parts are {\bf not} the same as those obtained during the processing phase. More precisely, they are shifted by the number of descriptive features.

We remove the cases with no clear output. For instance, it is possible to have a violation of a certain aspect of a given article but no violation of another aspect of the same article. In the future, we will consider a lower label level than the article.

\section{Experiments}
\label{sec:experiments}

In this section, we perform a first campaign of experiments on each of the produced datasets. The goals are twofold: studying the predictability offered by those datasets and their different flavors, and providing a first baseline by testing the most popular machine learning algorithms for classification. All the experiments are implemented using Scikit-Learn (\cite{scikit-learn}). We split the experiments into three categories: binary, multiclass and multilabel classifications, mostly because the evaluation metrics and their interpretation differ. All the experiments and scripts to analyze the results and generate the plots and tables are open-source and available on a separated GitHub repository\footnote{\url{https://github.com/aquemy/ECHR-OD_predictions}} for replication.

\subsection{Binary classification}

We are interested in answering four questions: 1) what is the predictive power of the datasets, 2) are all the articles equal w.r.t. predictability, 3) are some methods performing significantly better than others, and 3) are all dataset flavors equal w.r.t. predictability?  

\subsubsection{Protocol}

We compared 13 standard classification methods: AdaBoost with Decision Tree, Bagging with Decision Tree, Naive Bayes (Bernoulli and Multinomial), Decision Tree, Ensemble Extra Tree, Extra Tree, Gradient Boosting, K-Neighbors, SVM (linear, RBF), Neural Network (Multilayer Perceptron) and Random Forest.

For each article, we used three flavors: descriptive features only, bag-of-words only, and descriptive features combined to bag-of-words. For each method, each article and each flavor, we performed a 10-fold cross-validation with stratified sample, for a total of 429 validation procedures. Due to this important amount of experimental settings, we discarded the TF-IDF flavors. For the same reason, we did not perform any hyperparameter tuning at this stage.


To evaluate the performances, we reported some standard performance indicators: accuracy, $F_1$-score and Matthews correlation coefficient (MCC). Denoting by TP the number of true positives, TN the true negatives, FP the false positives and FN the false negative, those metrics are defined by:
\begin{align*}
\text{ACC} & = \frac{\text{TP} + \text{TN}}{\text{TP} + \text{TN} + \text{FP} + \text{FN}} \\
&\\
\text{F}_1 & = \frac{2 \text{TP}}{ 2\text{TP} + \text{FP} + \text{FN}} \\
    & \\
\text{MCC} & = \frac{\text{TP} \times \text{TN} - \text{FP} \times \text{FN}}{\sqrt{(\text{TP} + \text{FP})(\text{TP} + \text{FN})(\text{TN} + \text{FP})(\text{TN} + \text{FN})}}
\end{align*}
The accuracy, $\text{F}_1$-score and MCC respectively belongs to $[0,1]$, $[0,1]$ and $[-1,1]$. The closer to 1, the better it is. $\text{F}_1$-score and MCC take into account false positive and false negatives. Furthermore, MCC has been shown to be more informative than other metrics derived from the confusion matrix \cite{Chicco2017}, in particular with imbalanced datasets.

Additionally, we report the learning curves to study the limit of the model space for each method. The learning curves are obtained by plotting the accuracy depending on the training set size, for both the training and the test sets. The learning curves help to understand if a model underfit or overfit and thus, shape future axis of improvements to build better classifiers.

To find out what type of features are the most important w.r.t. predictability, we used a Wilcoxon signed-rank test at 5\% to compare the accuracy obtained on Bag-of-Words representation to the one obtained on the Bag-of-Words combined with the descriptive features. Wilcoxon signed-rank test is a non-parametric paired difference test. Given two paired sampled, the null hypothesis assumes the difference between the pairs follows a symmetric distribution around zero. The test is used to determine if the changes in the accuracy is significant when the descriptive features are added to the textual features.

\subsubsection{Results}

\begin{table}
  \caption{Best accuracy obtained for each article.}
  \label{table:summary_acc}
  \centering
  \begin{tabular}{|l|l|l|l| }
\hline
Article & Accuracy & Method & Flavor \\ \hline
Article 1 & 0.9832 (0.01) & Linear SVC & Descriptive features and Bag-of-Words\\
Article 2 & 0.9760 (0.02) & Linear SVC & Descriptive features and Bag-of-Words\\
Article 3 & 0.9588 (0.01) & BaggingClassifier & Descriptive features and Bag-of-Words\\
Article 5 & 0.9651 (0.01) & Gradient Boosting & Descriptive features and Bag-of-Words\\
Article 6 & 0.9721 (0.01) & Linear SVC & Descriptive features and Bag-of-Words\\
Article 8 & 0.9542 (0.03) & Gradient Boosting & Descriptive features and Bag-of-Words\\
Article 10 & 0.9392 (0.04) & Ensemble Extra Tree & Bag-of-Words only\\
Article 11 & 0.9671 (0.03) & Ensemble Extra Tree & Descriptive features and Bag-of-Words\\
Article 13 & 0.9450 (0.02) & Linear SVC & Descriptive features only\\
Article 34 & 0.7586 (0.09) & AdaBoost & Descriptive features only\\
Article p1 & 0.9685 (0.02) & Gradient Boosting & Descriptive features and Bag-of-Words\\
Average & 0.9443 & & \\
Micro average & 0.9644 & & \\
\hline
\end{tabular}
\end{table}

Table \ref{table:summary_acc} shows the best accuracy obtained for each article as well as the method and the flavor of the dataset. For all articles, the best accuracy obtained is higher than the prevalence. The method performing the best is linear SVM, obtaining the best results on 4 out of 11 articles. Gradient Boosting accounts for 3 out 11 articles and Ensemble Extra Tree accounts for 2 articles. The standard deviation is rather low with 1\% up to 4\%, at the exception of article 34 with 9\%. The accuracy ranges from 75.86\% to 98.32\% with an average of 94.43\%. The micro-average that ponders each result by the dataset size is 96.44\%. In general, the datasets with higher accuracy are larger and more imbalanced. The datasets being highly imbalanced, with a prevalence from 0.64 to 0.93, other metrics may be more suitable to appreciate the quality of the results. In particular, the micro-average could simply be higher due to the class imbalance rather than the availability of data.

Regarding the flavor, 8 out 10 best results are obtained on descriptive features combined to bag-of-words. Bag-of-words only is the best flavor for article 10 and descriptive features only for article 13 and article 34. This seems to indicate that combining information from different sources are improving the overall results.

\begin{center}
\begin{figure}
   \caption{\label{fig:normalized_cm} Normalized Confusion Matrices for the best methods as described by Table \ref{table:summary_acc}.}
   \includegraphics[scale=0.3]{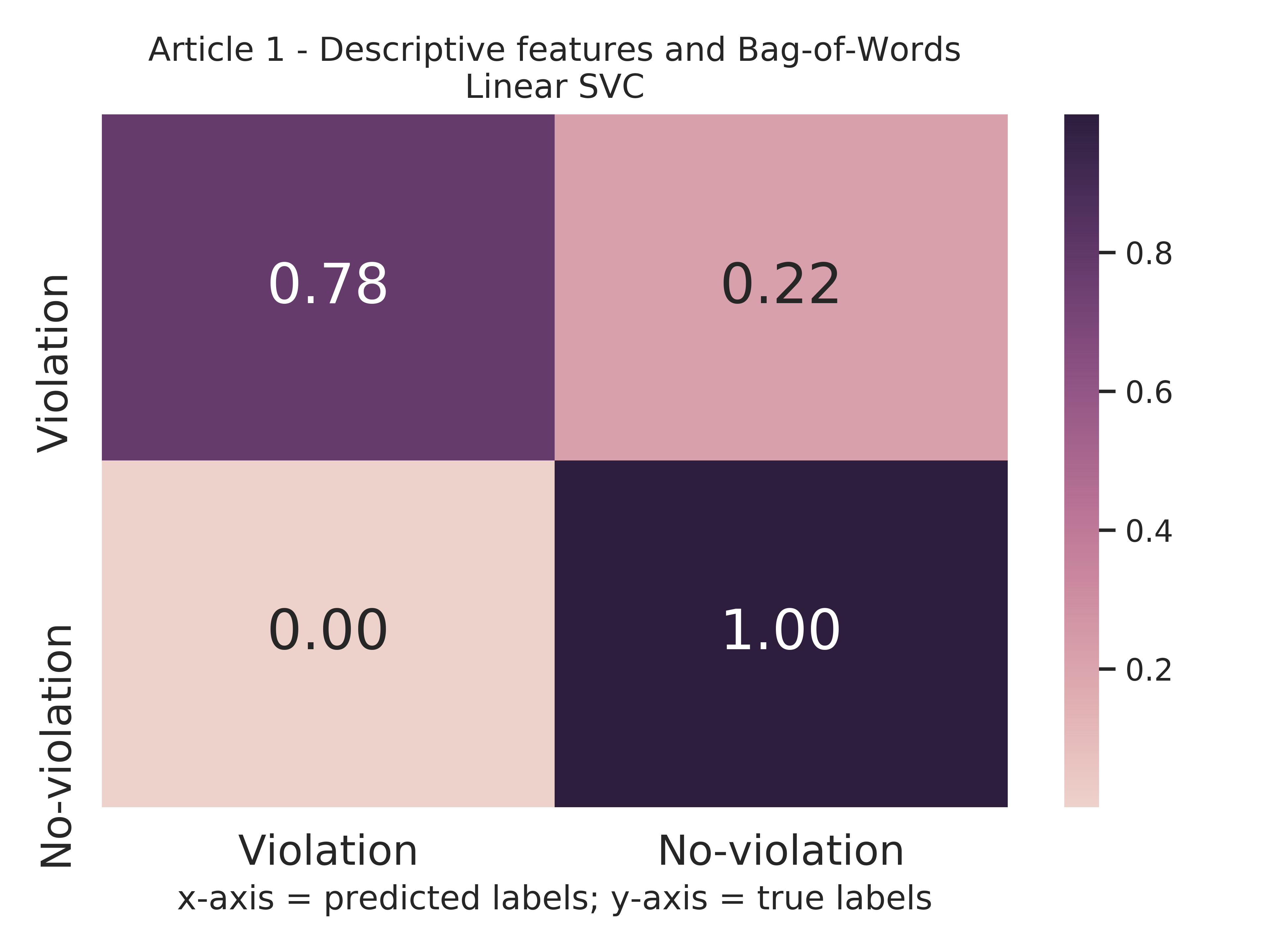}
   \includegraphics[scale=0.3]{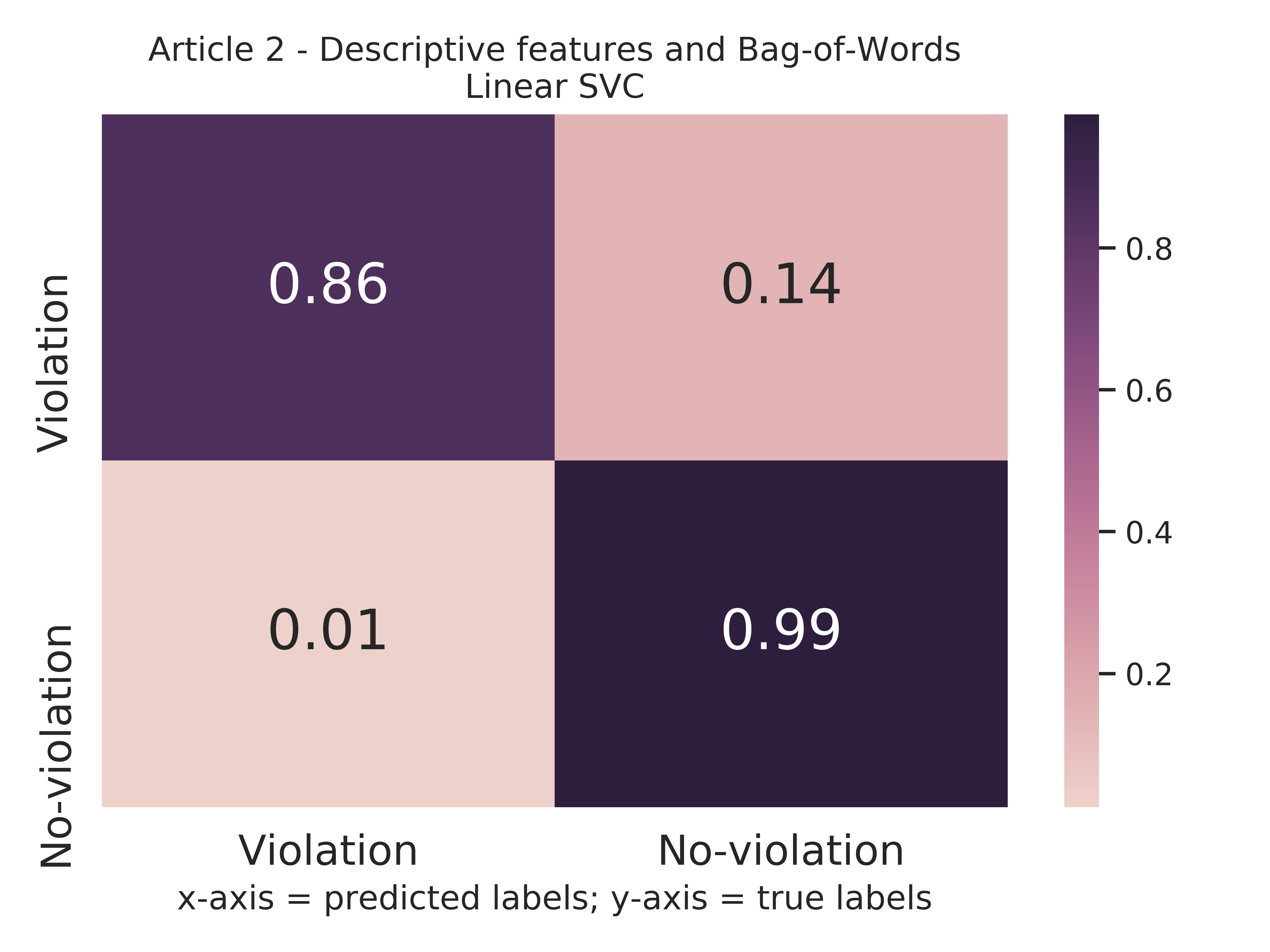}
   \includegraphics[scale=0.3]{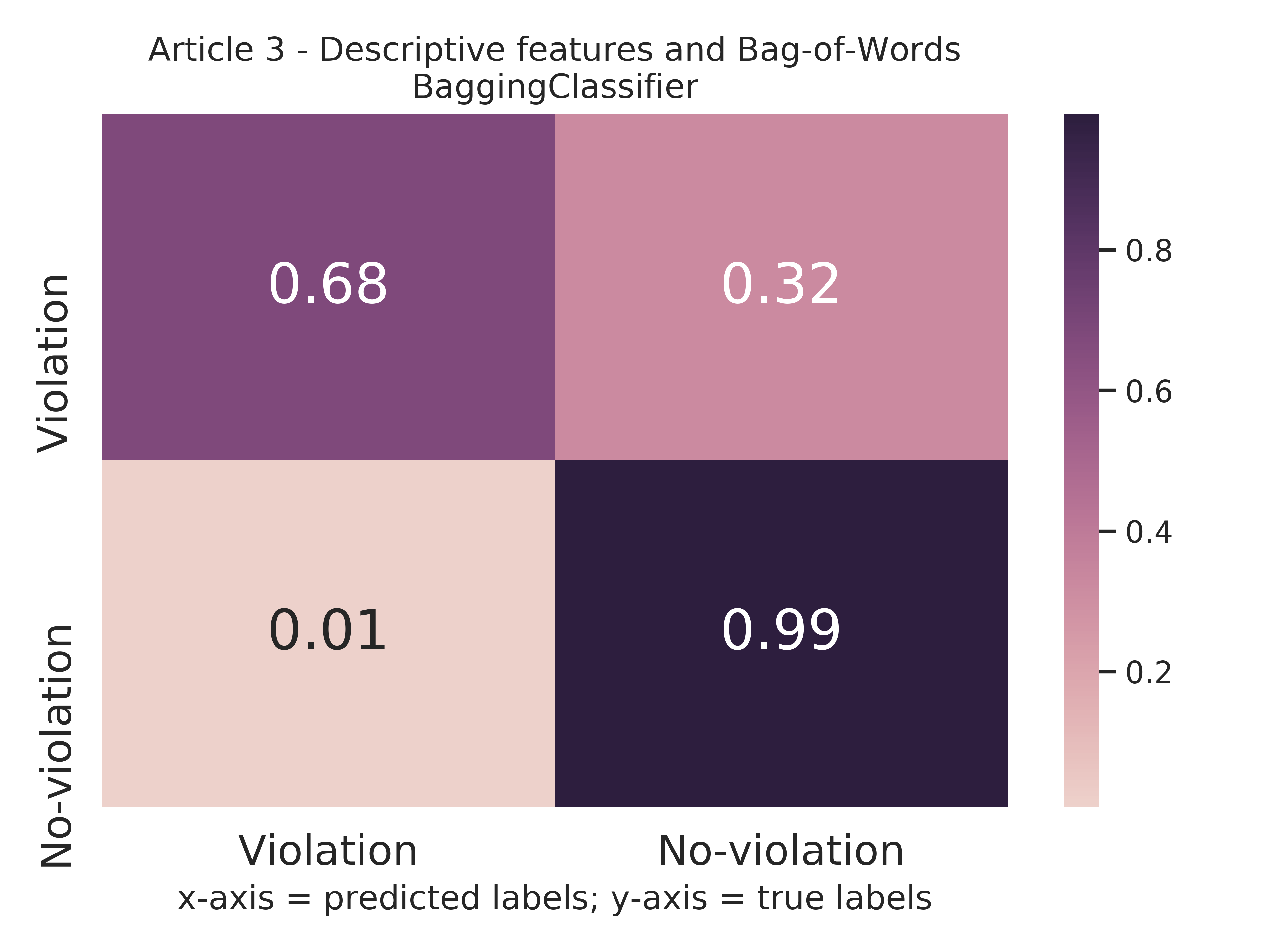}
   \includegraphics[scale=0.3]{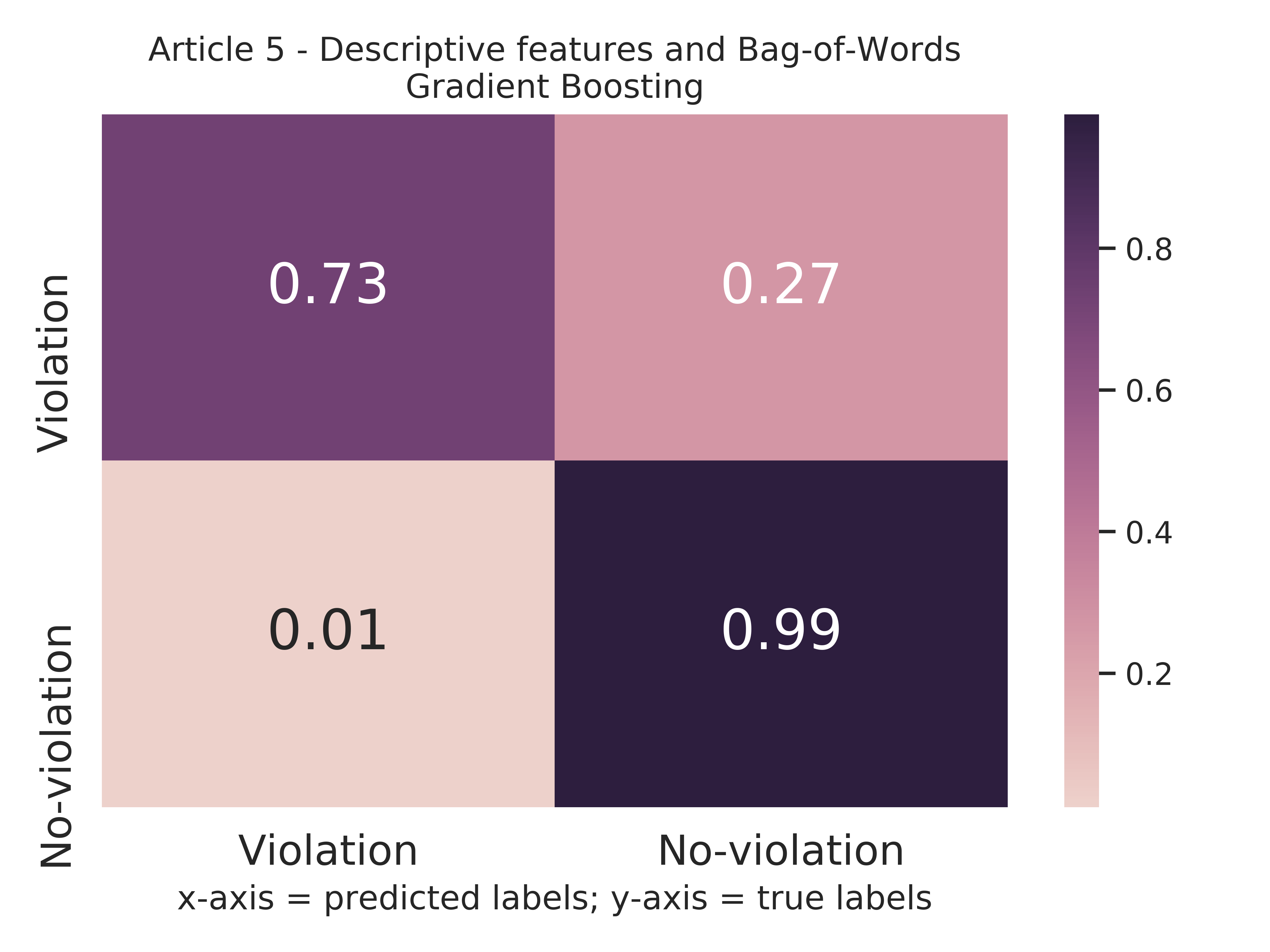}
   \includegraphics[scale=0.3]{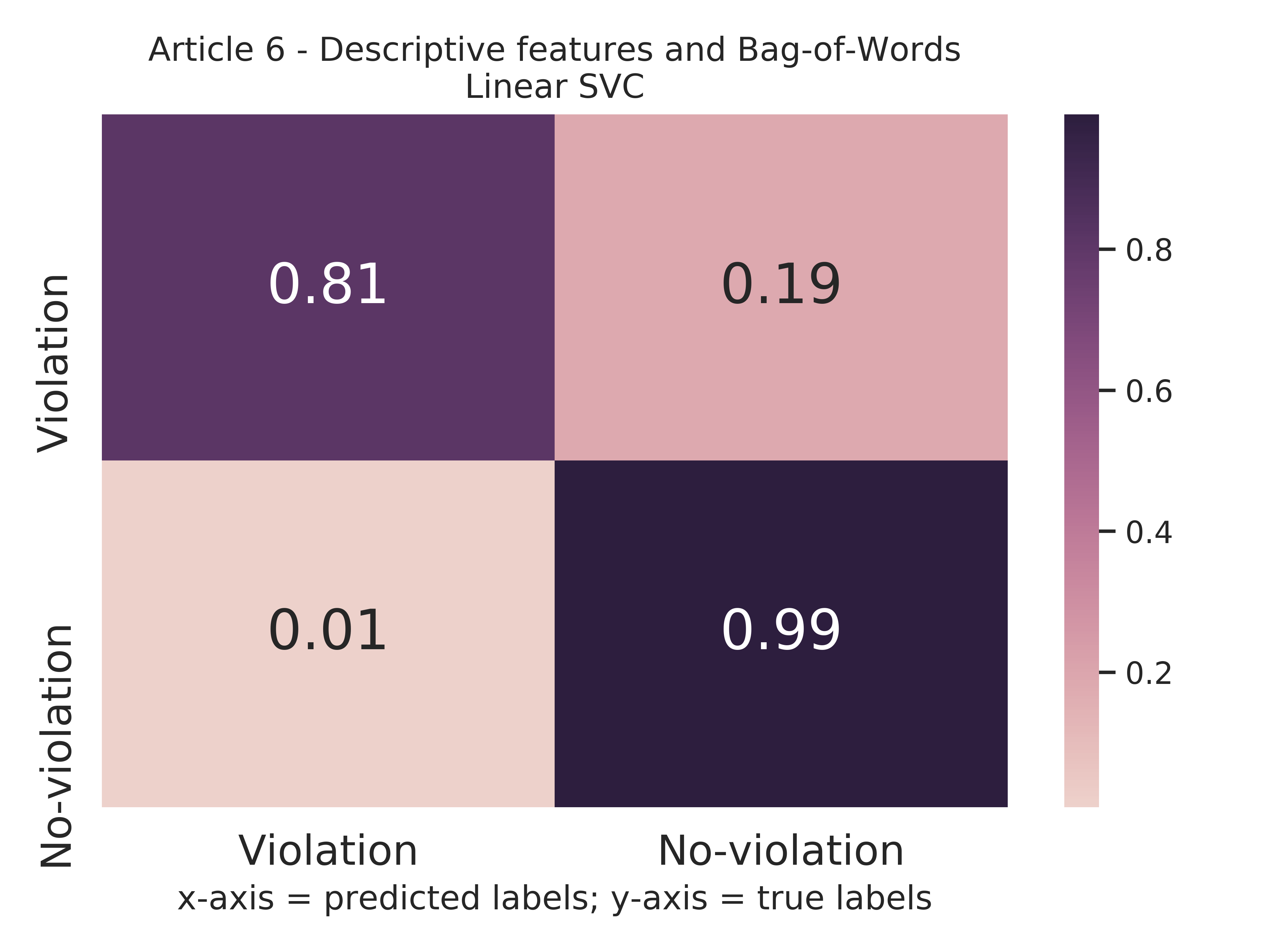}
   \includegraphics[scale=0.3]{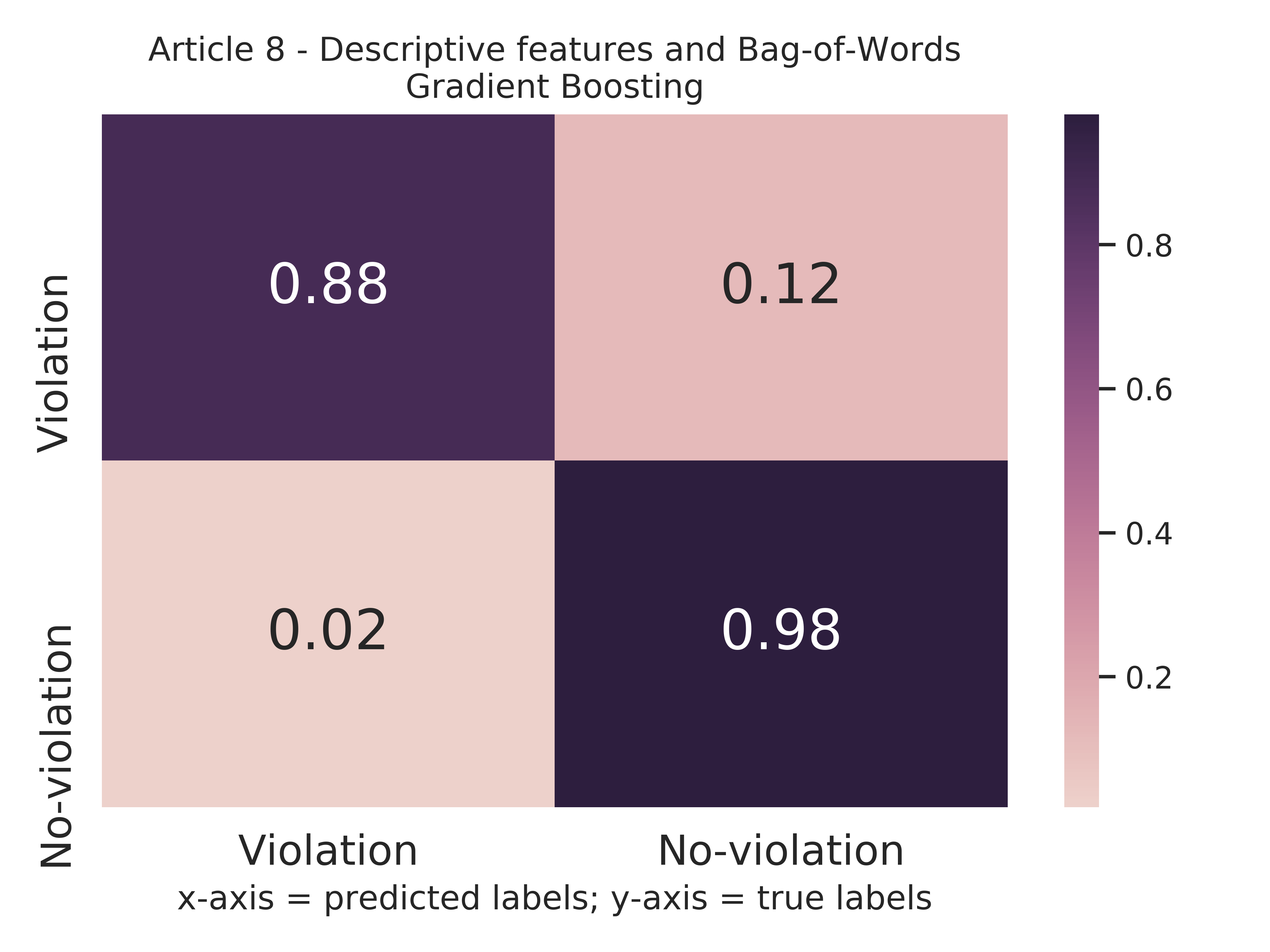}
   \includegraphics[scale=0.3]{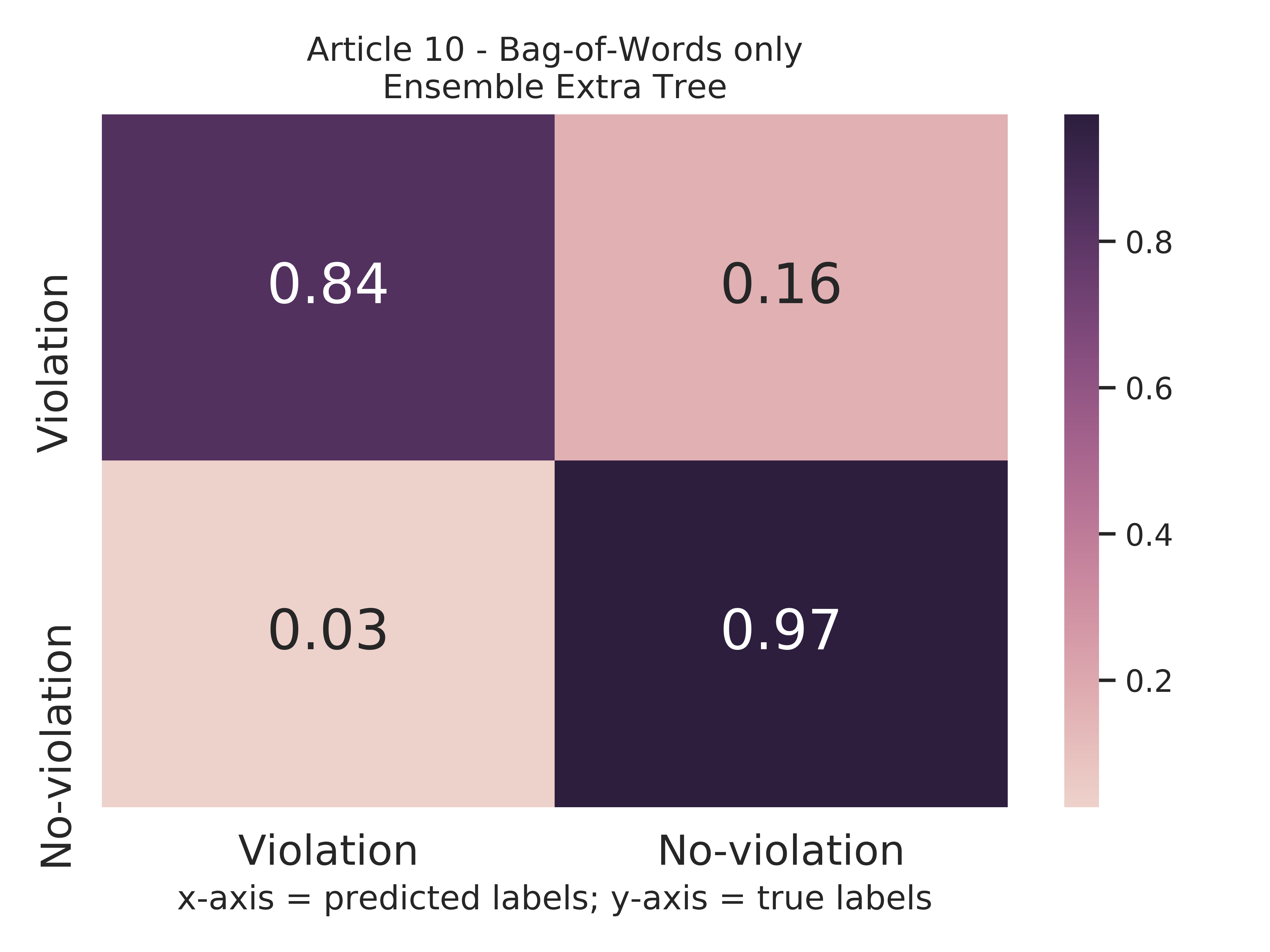}
   \includegraphics[scale=0.3]{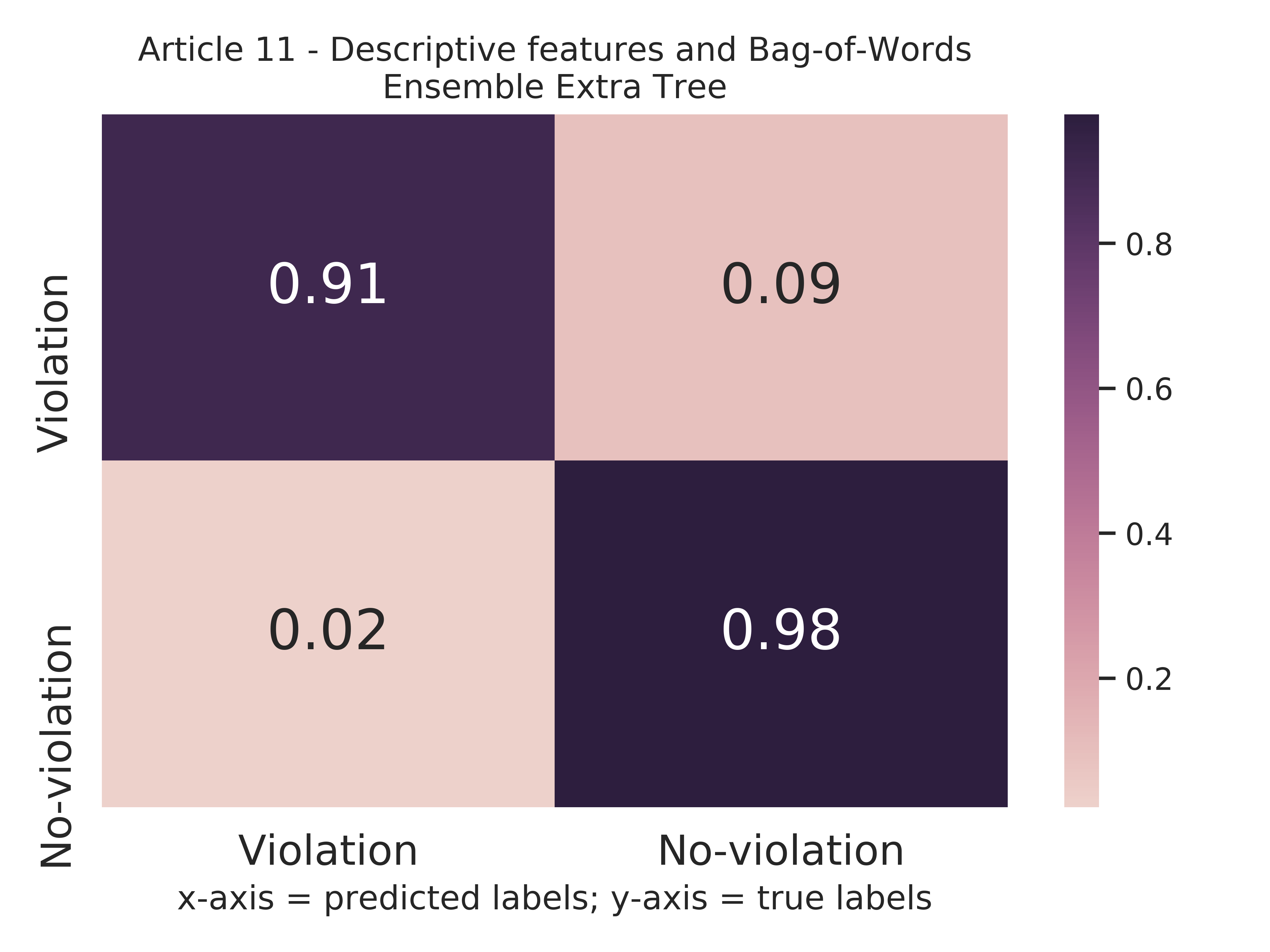}
   \includegraphics[scale=0.3]{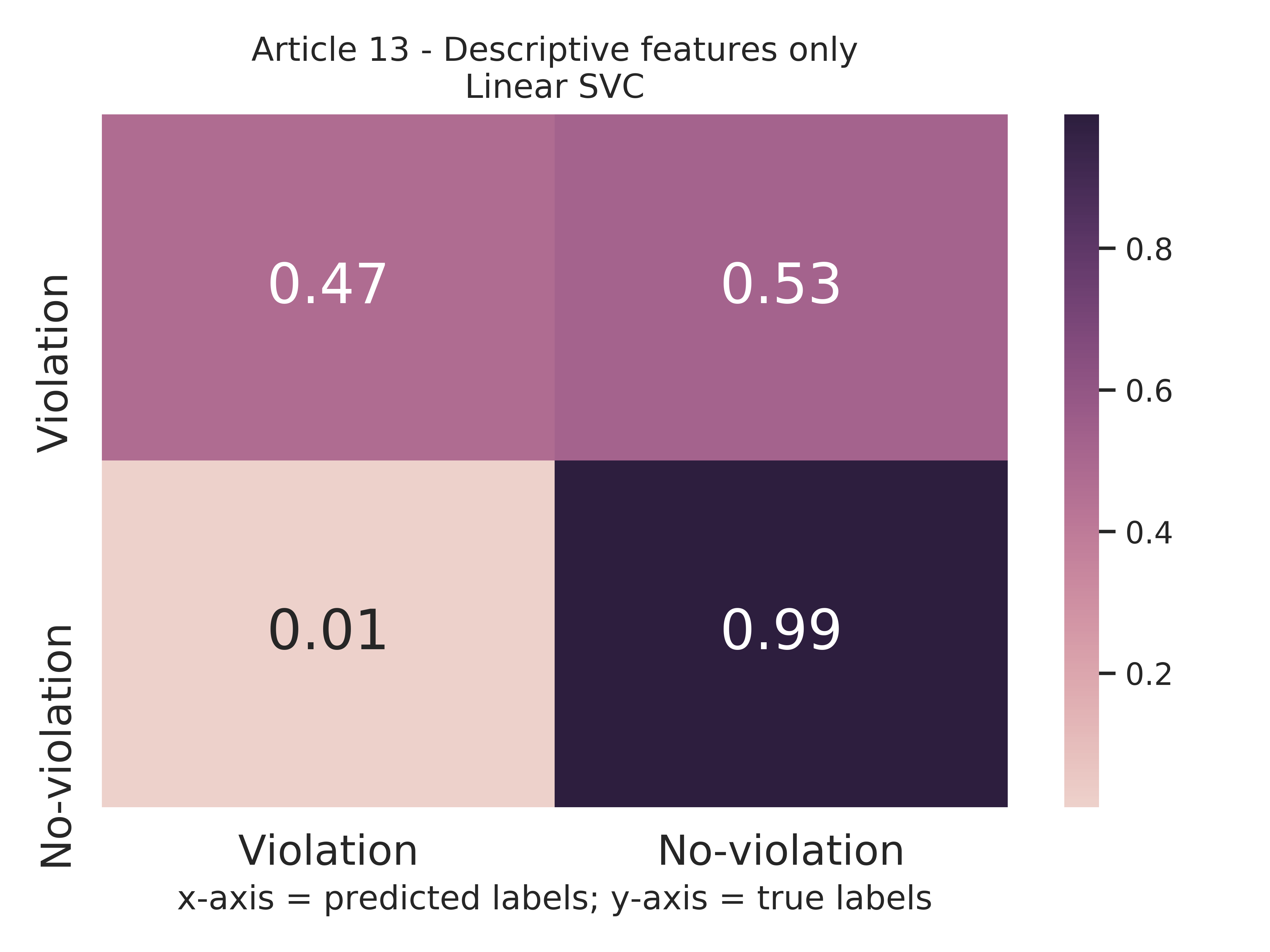}
   \includegraphics[scale=0.3]{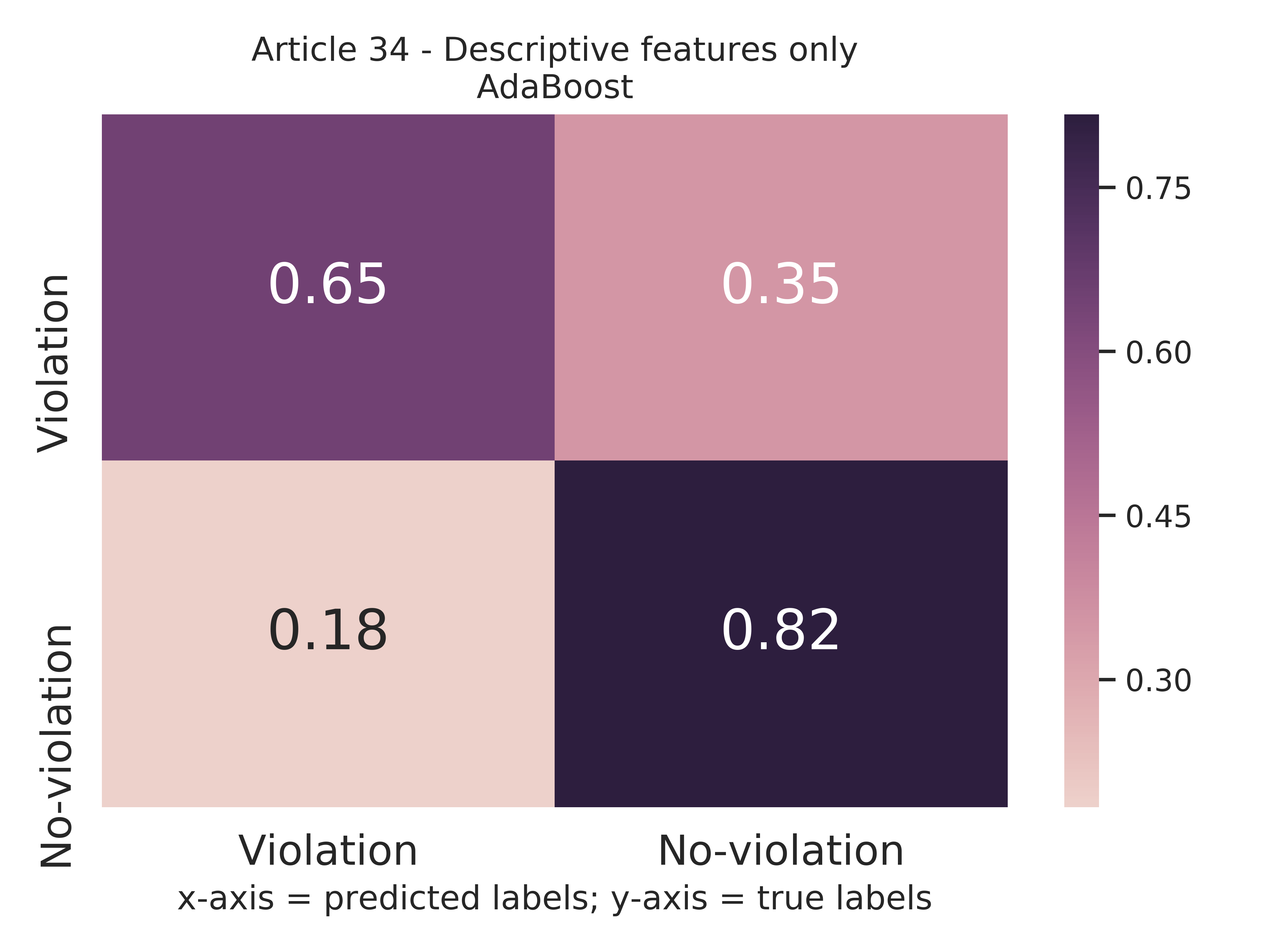}
   \includegraphics[scale=0.3]{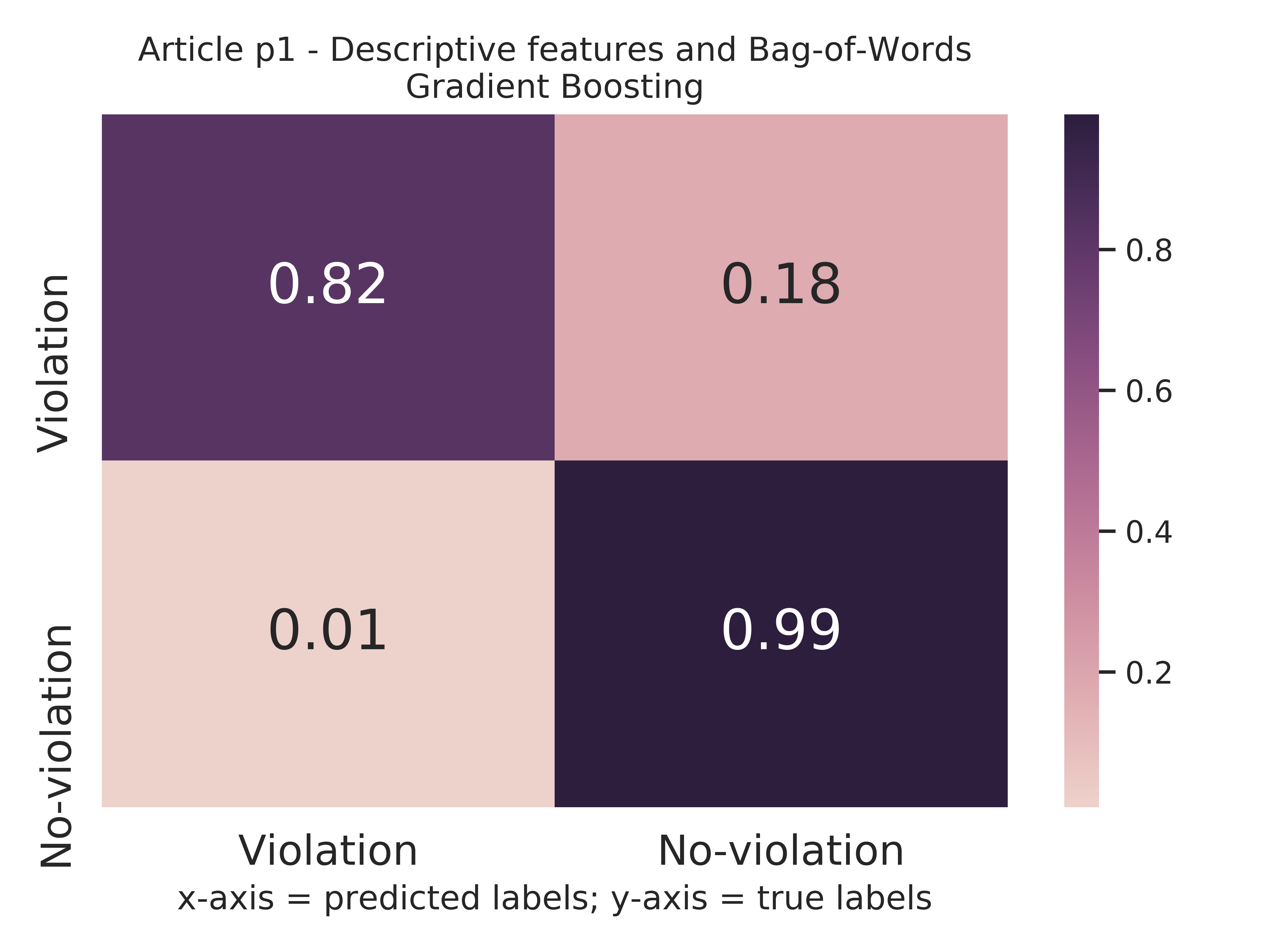}
\end{figure}
\end{center}

Figure \ref{fig:normalized_cm} displays the normalized confusion matrix for each line of the Table \ref{table:summary_acc}. The normalization is done per line and allows to quickly appreciate how the true predictions are balanced for both classes. As expected by the prevalence, true negatives are extremely high, ranging from 0.82 to 1.00 with an average of 97.18. On the contrary, the true positive rate is lower ranging from 0.47 to 0.91. For most articles, the true positive rate is higher than 80\% and is lower than 50\% only for article 34.

\begin{table}
  \caption{Best Matthews Correlation Coefficient and F1 score obtained for each article. The flavor and method achieving the best score for both metrics are similar for every article.}
  \label{table:summary_mcc}
  \centering
  \begin{tabular}{|l|l|l|l| }
\hline
Article & MCC & Method & Flavor \\ \hline
Article 1 & 0.8654 & Linear SVC & Descriptive features and Bag-of-Words\\
Article 2 & 0.8609 & Linear SVC & Descriptive features and Bag-of-Words\\
Article 3 & 0.7714 & BaggingClassifier & Descriptive features and Bag-of-Words\\
Article 5 & 0.7824 & Gradient Boosting & Descriptive features and Bag-of-Words\\
Article 6 & 0.8488 & Linear SVC & Descriptive features and Bag-of-Words\\
Article 8 & 0.8829 & Gradient Boosting & Descriptive features and Bag-of-Words\\
Article 10 & 0.8411 & Gradient Boosting & Bag-of-Words only\\
Article 11 & 0.8801 & Ensemble Extra Tree & Descriptive features and Bag-of-Words\\
Article 13 & 0.5770 & Ensemble Extra Tree & Bag-of-Words only\\
Article 34 & 0.4918 & AdaBoost & Descriptive features only\\
Article p1 & 0.8656 & Gradient Boosting & Descriptive features and Bag-of-Words\\
Average & 0.7879 & & \\
Micro average & 0.8163 & & \\
\hline
\end{tabular}
\end{table}

In addition, we provide the Matthew Correlation Coefficient in Table \ref{table:summary_mcc} and the F$_1$-score in Supplementary Material. The F1-score is weighted by the class support to account for class imbalance. The Matthew Correlation Coefficient ranges from 0.4918 on article 34 to 0.8829 on article 10. The best score is not obtained by the same article as for accuracy (article 10 achieved 93\% accuracy, below the average). Interestingly, the MCC reveals that the performances on article 34 are rather poor in comparison to the other articles and close to those of article 13. 
Surprisingly, the best method is not linear SVC anymore (best on 3 articles) but Gradient Boosting (best on 4 articles). 
While the descriptive features were returning the best results for two articles, according the Matthew Correlation Coefficient, it reaches the best score only for article 34. 

Once again, the micro-average is higher than the macro-average. As the MCC and the weighted F$_1$-score take into account class imbalance, it supports the idea that adding more cases to the training set could still improve the result of those classifiers. This will be confirmed by looking at the learning curves.

Table \ref{table:summary_methods_accuracy} ranks the methods according to the average accuracy performed on all articles. For each article and method, we kept only the best accuracy among the three dataset flavors. Surprisingly, Linear SVC or Gradient Boosting are not the best method with a respective rank of 2 and 5, but Ensemble Extra Tree. Random Forest and Bagging with Decision Tree are respectively second and third while they never achieved the best result on any article. It simply indicates that those methods are more consistent across the datasets than Linear SVM or Gradient Boosting. This can be confirmed by the detailed results per article provided in the Supplementary Material.

\begin{table}
  \caption{Overall ranking of methods according to the average accuracy obtained on every article.}
  \label{table:summary_methods_accuracy}
  \centering
  \begin{tabular}{|l|l|l|l| }
\hline
Method                  & Accuracy & Micro Accuracy & Rank \\ \hline
Ensemble Extra Tree     & 0.9420 & 0.9627 & 1\\
Linear SVC              & 0.9390 & 0.9618 & 2\\
Random Forest           & 0.9376 & 0.9618 & 3\\
BaggingClassifier       & 0.9319 & 0.9599 & 4\\
Gradient Boosting       & 0.9309 & 0.9609 & 5\\
AdaBoost                & 0.9284 & 0.9488 & 6\\
Neural Net              & 0.9273 & 0.9535 & 7\\
Decision Tree           & 0.9181 & 0.9419 & 8\\
Extra Tree              & 0.8995 & 0.9275 & 9\\
Multinomial Naive Bayes & 0.8743 & 0.8907 & 10\\
Bernoulli Naive Bayes   & 0.8734 & 0.8891 & 11\\
K-Neighbors             & 0.8670 & 0.8997 & 12\\
RBF SVC                 & 0.8419 & 0.8778 & 13\\
Average & 0.9086 & 0.9335 & \\
\hline
\end{tabular}
\end{table}

\begin{center}
\begin{figure}
   \caption{\label{fig:learning_curves} Learning Curves for the best methods as described by Table \ref{table:summary_acc}.}
   \includegraphics[scale=0.3]{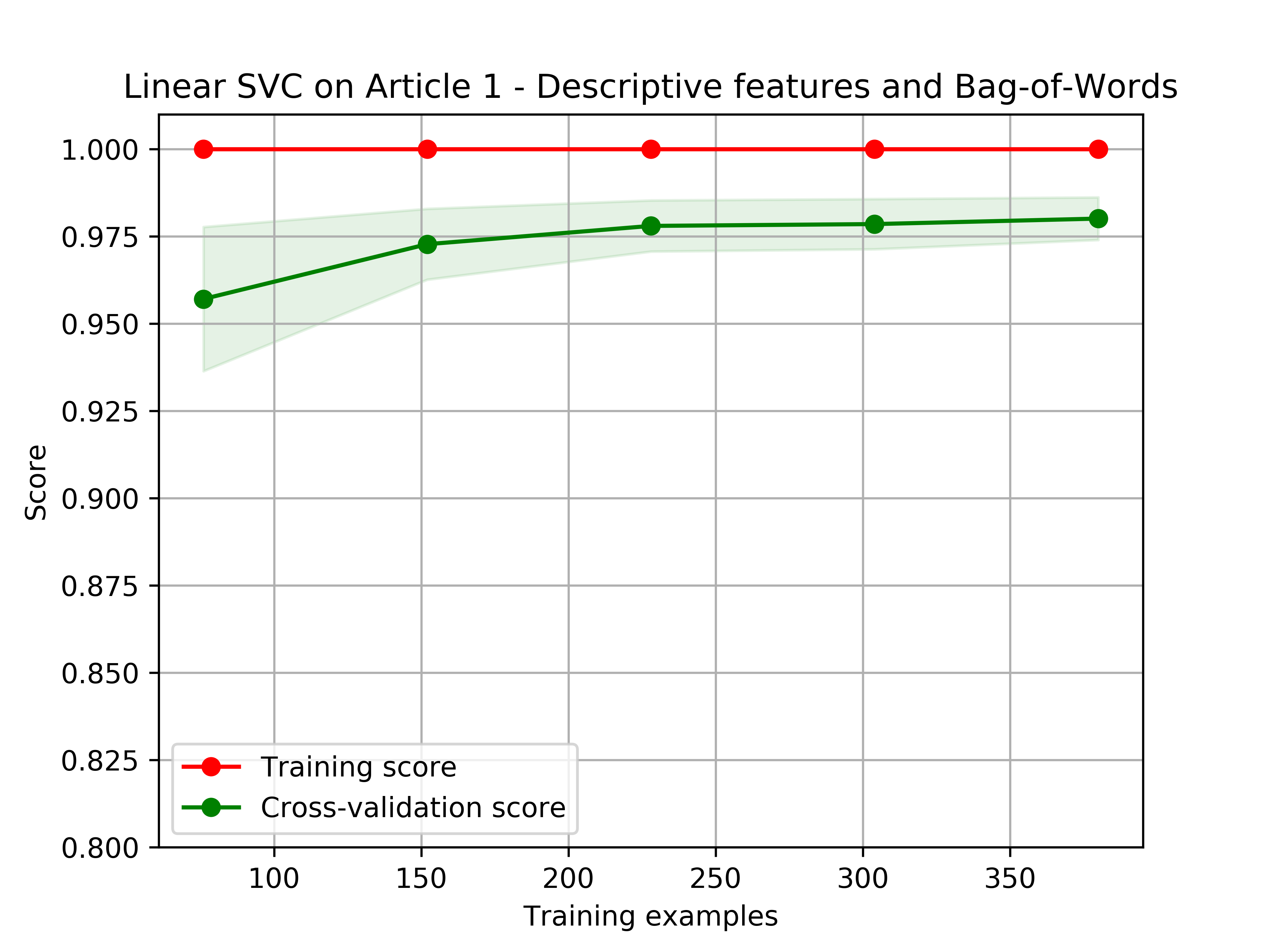}
   \includegraphics[scale=0.3]{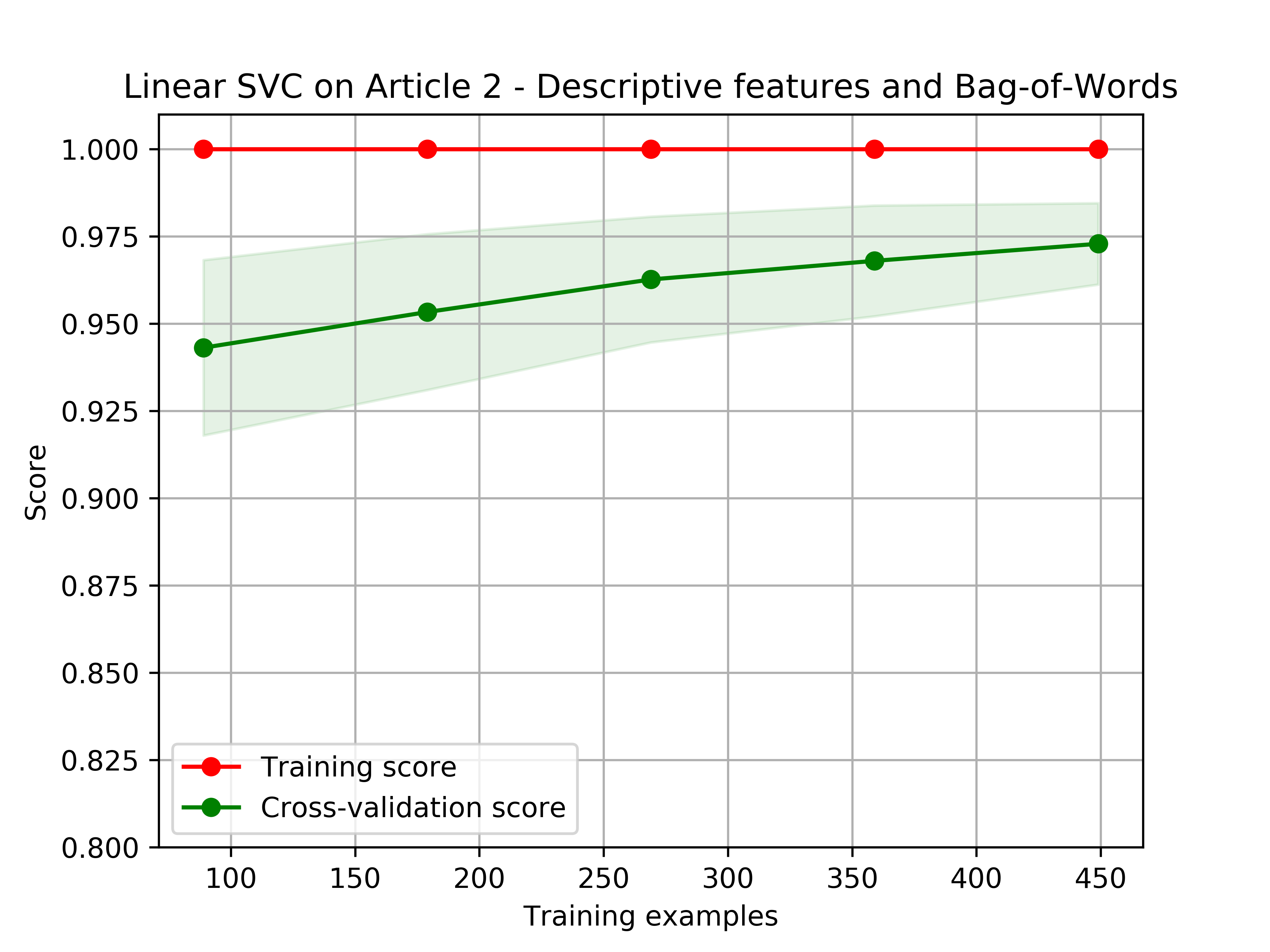}
   \includegraphics[scale=0.3]{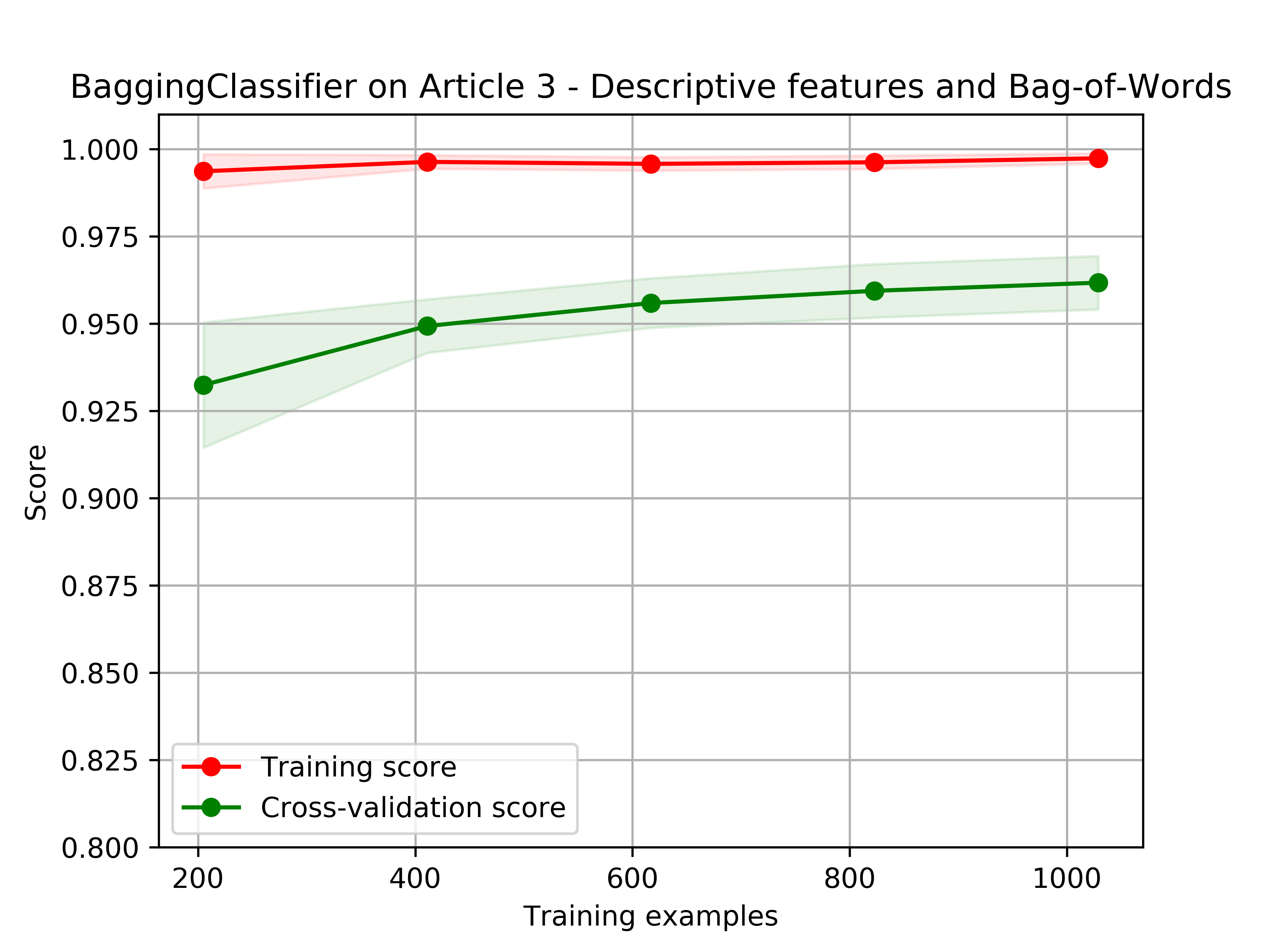}
   \includegraphics[scale=0.3]{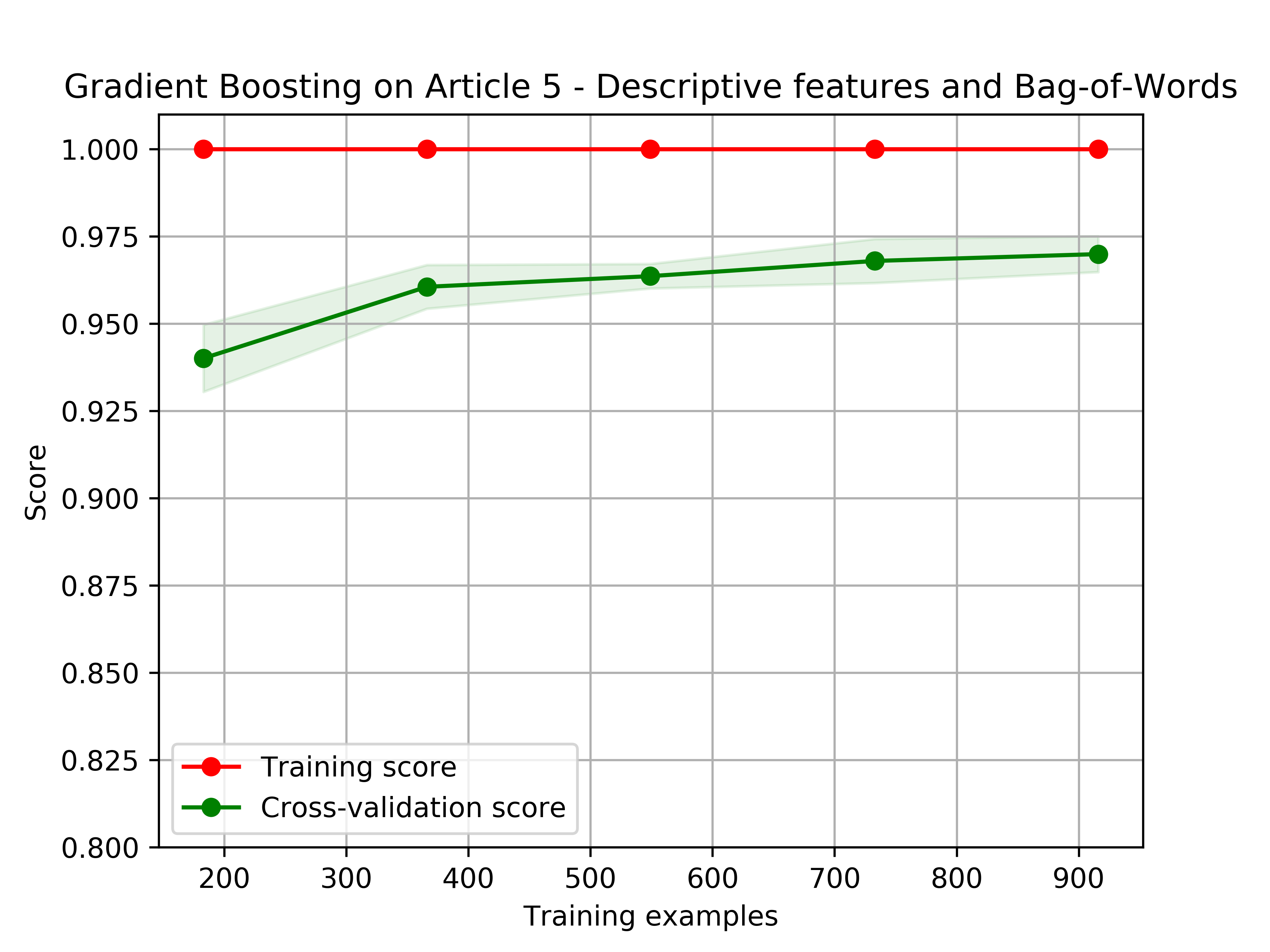}
   \includegraphics[scale=0.3]{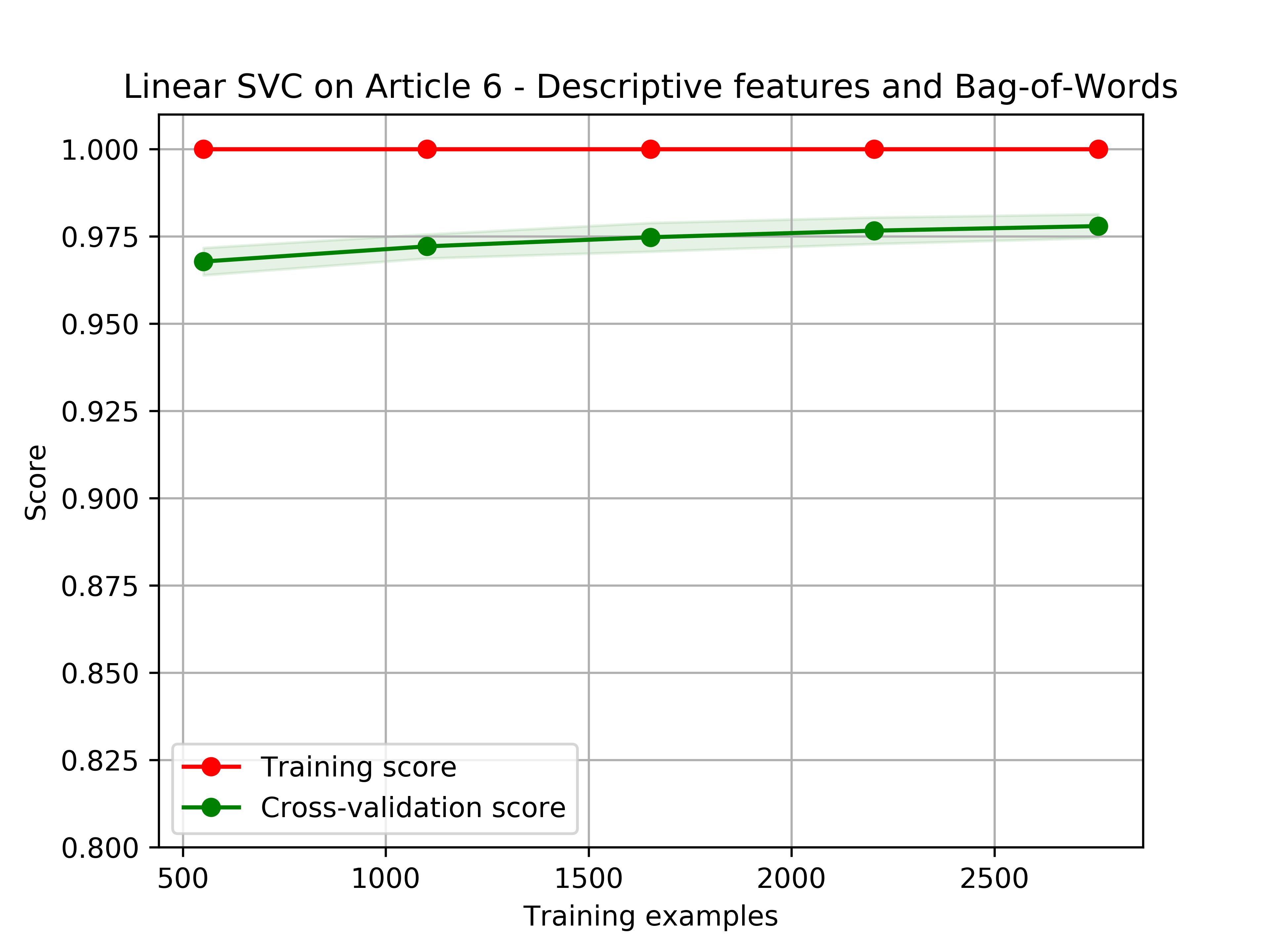}
   \includegraphics[scale=0.3]{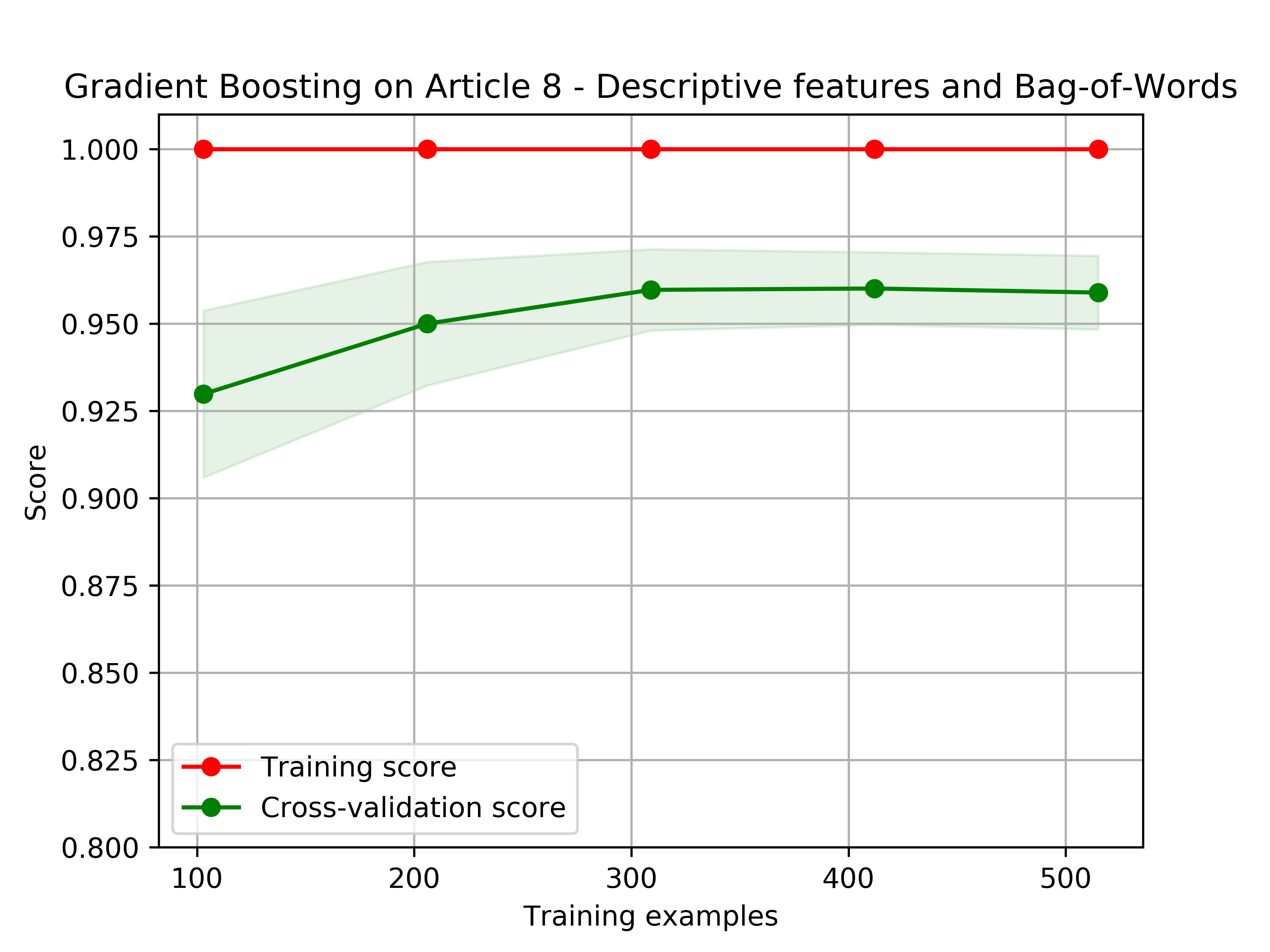}
   \includegraphics[scale=0.3]{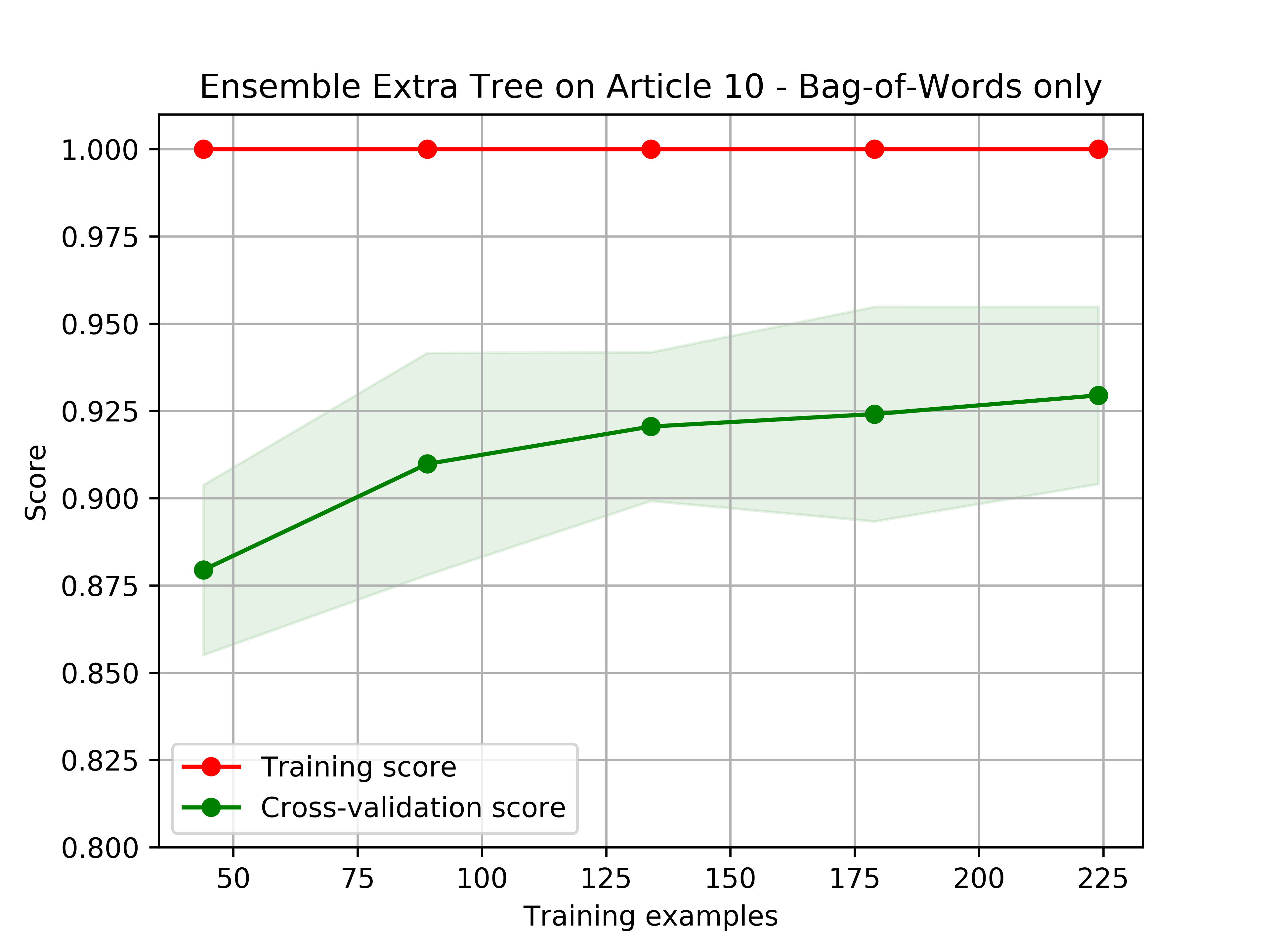}
   \includegraphics[scale=0.3]{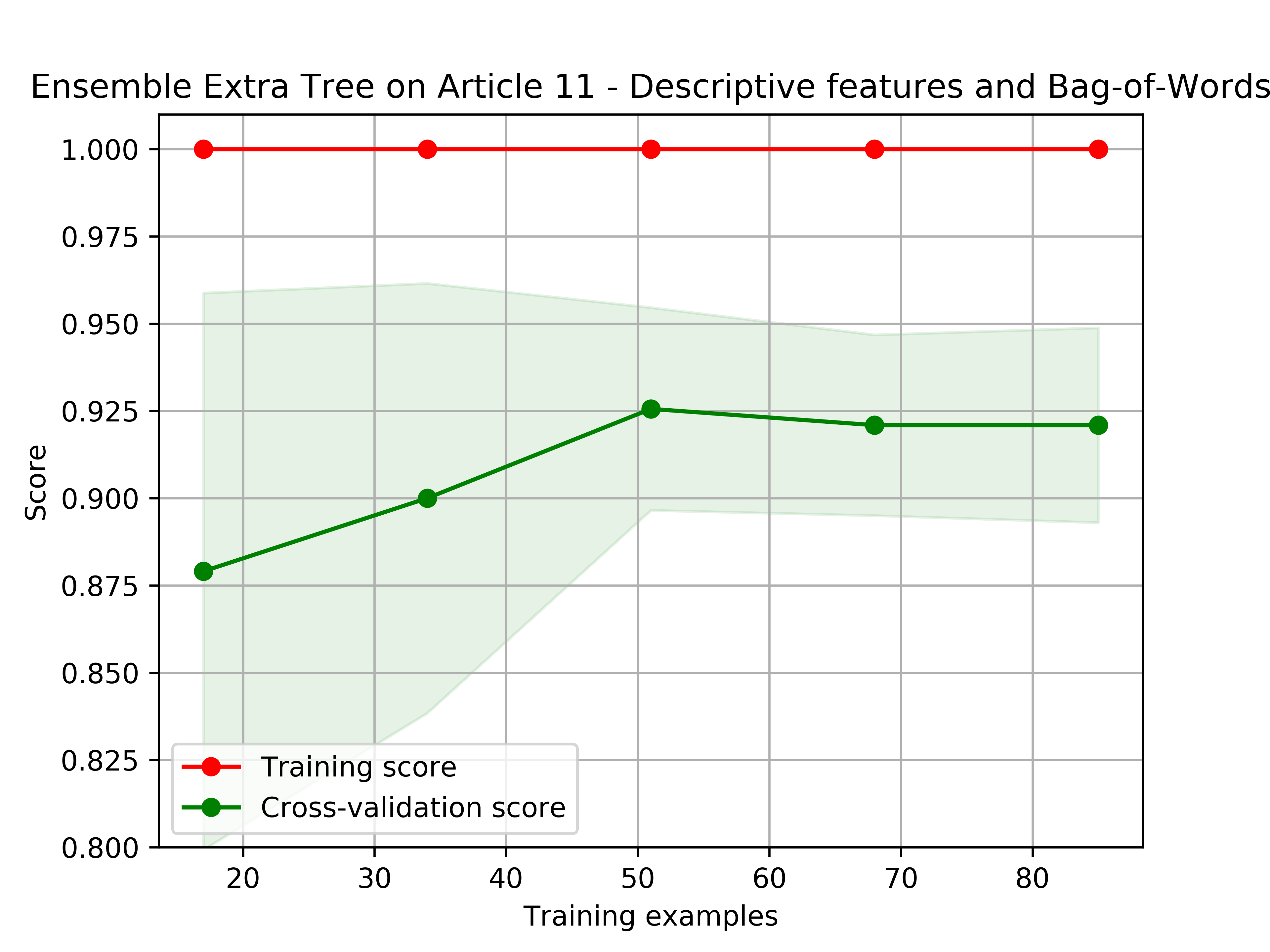}
   \includegraphics[scale=0.3]{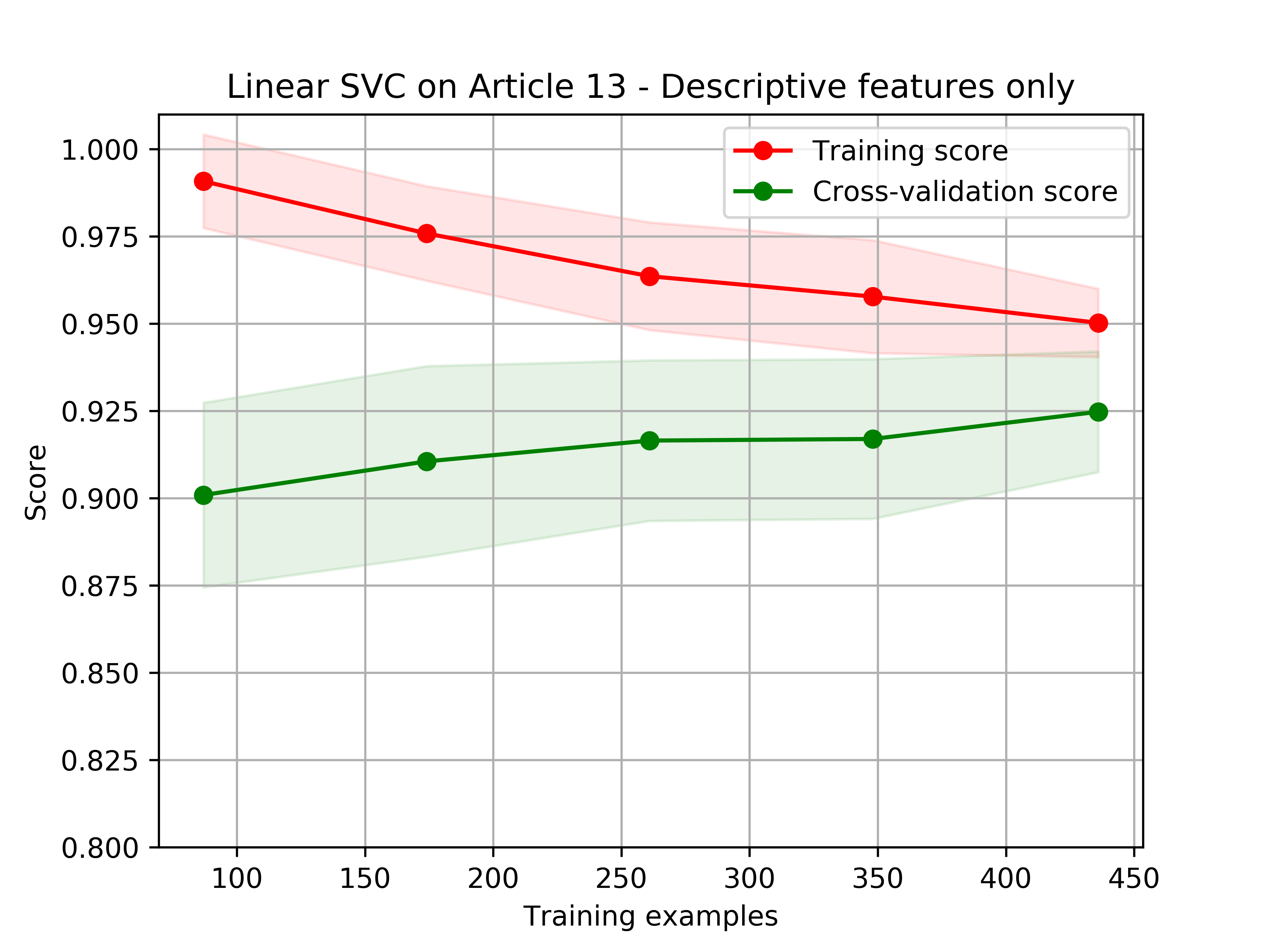}
   \includegraphics[scale=0.3]{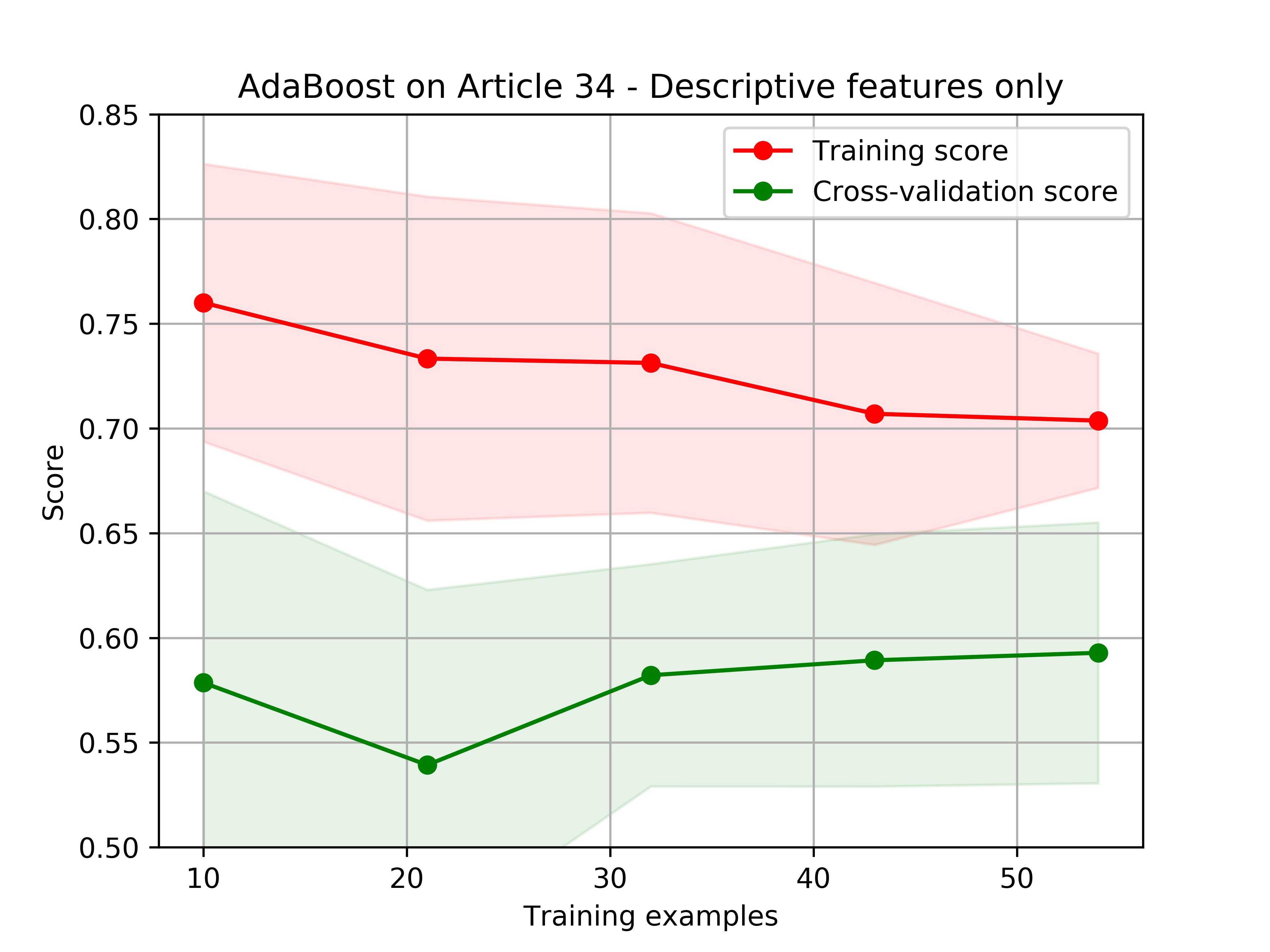}
   \includegraphics[scale=0.3]{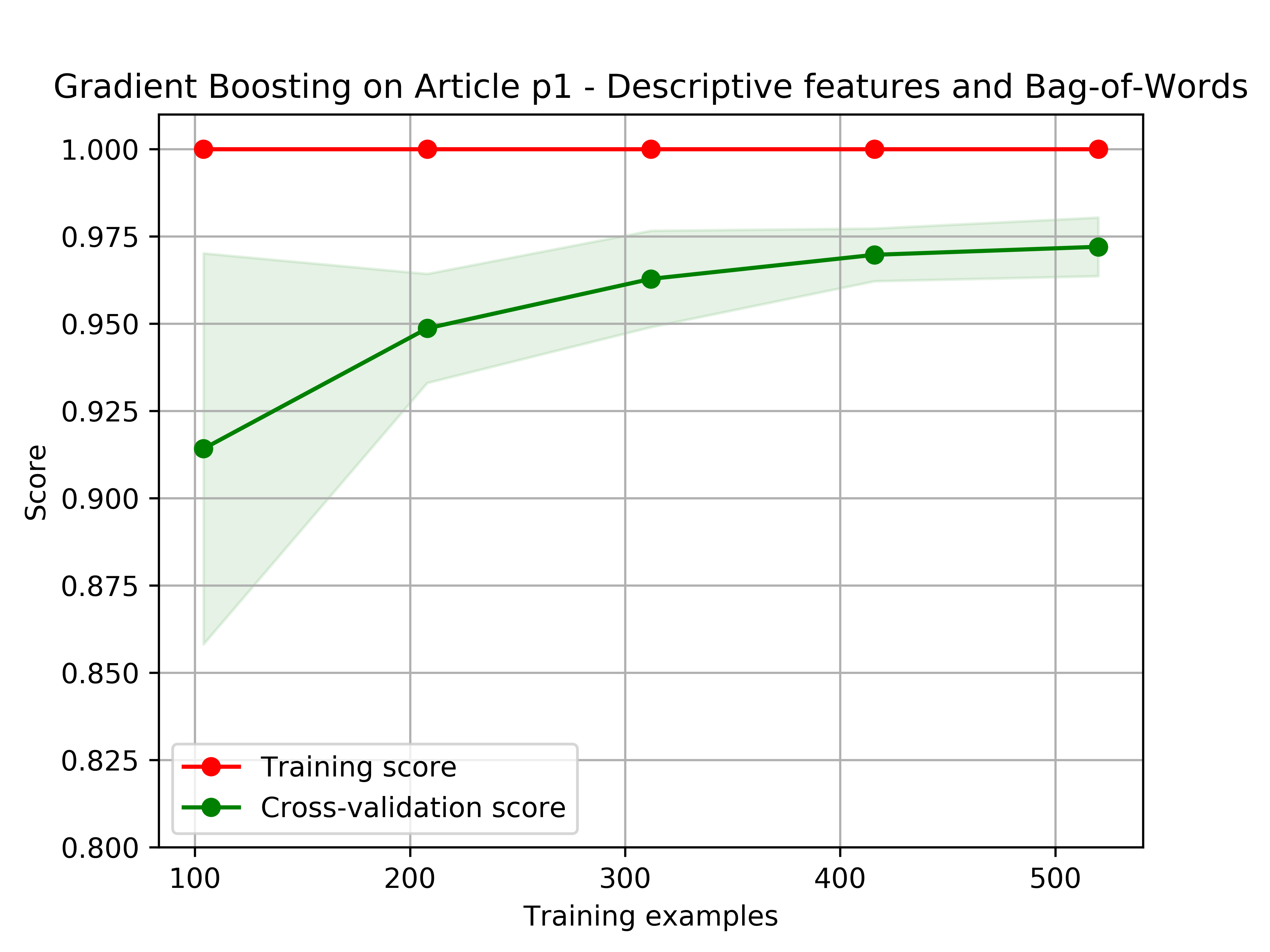}
\end{figure}
\end{center}

Figure \ref{fig:learning_curves} displays the learning curves obtained for the best methods described by Table \ref{table:summary_acc}. The training error becomes (near) zero on every instance after only few cases, except for article 13 and 34. The test error converges rather fast and remains relatively far from the training error, synonym of high bias. Those two elements indicate underfitting. Usually, more training examples would help but as the datasets are exhaustive w.r.t. the European Court of Human Rights cases, this is not possible. As a result, simpler model space has to be investigated as well as, in general, hyperparameter tuning. An exploratory analysis of the datasets may also help in removing some noise and finding the best predictors.

On article 13 and 34, the bias is also high, and variance relatively higher than for the other articles, clearly indicating the worst possible case. Again, adding more examples is not an option. However, if we assume that the process of deciding if there is a violation or not is the same independently of the article, a solution might be transfer learning to leverage what is learnable from the other articles. We let this research axis for future work.

Finally, we used a Wilcoxon signed-rank test at 5\% to compare the accuracy obtained on Bag-of-Words representation to the one obtained on the Bag-of-Words combined with the descriptive features. The difference between the sample has been found to be significant only for article 6 and article 8. 
The best result of column \texttt{BoW} is improved in the column \texttt{both} for every article. However, statistically, for a given method, adding descriptive features does not improve the result. 

Additionally, we performed the test per method. The result is significant for any method. 

In conclusion, the datasets demonstrated a strong predictability power. Apart from article 13 and 34, each article seems to provide similar results independently of the relatively different prevalence. If the accuracy is rather high w.r.t. the prevalence, more informative metrics such as MCC and F$_1$ scores shows that there are still margins of improvements. Hyperparameter tuning is an obvious way to go, and this preliminary work have shown that good candidates for fine tuning are Ensemble Extra Tree, Linear SVM and Gradient Boosting.
This experimental campaign has demonstrated that the textual information provides better results than descriptive features alone, but the addition of those descriptive feature improve in general the final {\bf best} result. We emphasize the best (obtained among all methods) because for a given method and any article, adding the descriptive feature are not significantly improving the results. Another way of improving the results is to tune the different phases of the dataset generations. In particular, preliminary work in \cite{DBLP:conf/dolap/Quemy19} have shown that 5000 tokens and $4$-grams might not be enough to take the best out of the documents. It might seem surprising, but the justice language is codified and standardized in a way that $n$-grams for large $n$ might be good predictors for the outcome.

\subsection{Multiclass classification}

In this section, we are interested in quantifying the capacity of standard machine learning algorithms to deal with the multiclass dataset. In the previous section, we showed that most methods could obtain an accuracy higher than the dataset prevalence, and more generally, good evaluation metrics. Usually algorithms for binary classification adapt relatively well to multiclass problems, however, in the case of ECHR-OD, the labels come by pair (violation or no-violation of a given article) which may confuse the classifiers.
The experimental protocol being similar to the one of the previous section, we describe the results right after. For computational purposes, we dropped the two worst classifiers on the binary datasets, namely RBF SVM and KNN.

\begin{table}
  \caption{Accuracy obtained for each method on the multiclass dataset.}
  \label{table:mc_acc}
  \centering
  \begin{tabular}{|l|l|l|l| }
\hline
 &  \multicolumn{3}{c|}{ Accuracy - Multiclass} \\
\cline{2-4} & desc & BoW & both \\ \hline
AdaBoost                & 0.7789 (0.04) & 0.5451 (0.15) & 0.5720 (0.04)\\
BaggingClassifier       & 0.8998 (0.01) & 0.8794 (0.01) & {\bf 0.9499} (0.01)\\
Bernoulli Naive Bayes   & 0.5096 (0.01) & 0.7516 (0.01) & 0.7464 (0.01)\\
Decision Tree           & 0.8897 (0.02) & 0.8457 (0.01) & 0.9434 (0.01)\\
Ensemble Extra Tree     & 0.8788 (0.01) & 0.8904 (0.01) & 0.9195 (0.01)\\
Extra Tree              & 0.7776 (0.03) & 0.7164 (0.03) & 0.7458 (0.02)\\
Gradient Boosting       & 0.8911 (0.01) & 0.8904 (0.01) & 0.9494 (0.01)\\
Linear SVC              & {\bf 0.9141} (0.01) & {\bf 0.9136} (0.01) & 0.9420 (0.01)\\
Multinomial Naive Bayes & 0.7980 (0.01) & 0.7784 (0.01) & 0.7829 (0.01)\\
Neural Net              & 0.8813 (0.01) & 0.9072 (0.01) & 0.9231 (0.01)\\
Random Forest           & 0.8669 (0.01) & 0.8825 (0.01) & 0.9125 (0.01)\\
\hline
\end{tabular}
\end{table}

Table \ref{table:mc_acc} presents the accuracy obtained on the multiclass dataset. The best accuracy for descriptive features only and Bag-of-Words only is linear SVM with respectively 91.41\% and 91.36\% correctly labeled cases. This is aligned with the results obtained on binary datasets. However, the top score of 94.99\% is obtained by Bagging Classifier that only ranked 4th on binary datasets. In other words, SVM ranked first on two types of features individually, but the improvement of combining the features is lower than the one obtained by Bagging Classifier. The same can be observed with Gradient Boosting that outperforms SVM. Except from Ada Boost, the standard deviation is mostly lower than 1\%. 

 For most methods, the flavor Bag-of-words only scores better than descriptive features. This observation is reversed by looking at the Matthew Correlation Coefficient provided by Table \ref{table:mc_mcc}. For both indicators however, combining both types of features increases performances at the notable exceptions of Extra Tree, Multinomial Naive Bayes and Ada Boost with Decision Tree.

This highly contrasts with the binary setting for which descriptive features were quantitatively far below textual features, in particular the MCC indicator. For binary datasets, the flavor descriptive features only was mostly scoring below the bag-of-words only, for any article and any method (c.f. Supplementary Material).
On top of that, taking only the best result per flavor, the descriptive features score better than purely textual features. The explanation can be found by studying the confusion matrix.

\begin{table}
  \caption{Matthew Correlation Coefficient obtained for each method on the multiclass dataset.}
  \label{table:mc_mcc}
  \centering
  \begin{tabular}{|l|l|l|l| }
\hline
 &  \multicolumn{3}{c|}{ MCC - Multiclass} \\
\cline{2-4} & desc & BoW & both \\ \hline
AdaBoost                & 0.7171 (0.04) & 0.4416 (0.14) & 0.4580 (0.05)\\
BaggingClassifier       & 0.8700 (0.02) & 0.8435 (0.02) & {\bf 0.9353} (0.01)\\
Bernoulli Naive Bayes   & 0.2750 (0.02) & 0.6845 (0.01) & 0.6664 (0.01)\\
Decision Tree           & 0.8572 (0.02) & 0.8004 (0.02) & 0.9268 (0.01)\\
Ensemble Extra Tree     & 0.8419 (0.01) & 0.8576 (0.01) & 0.8956 (0.01)\\
Extra Tree              & 0.7103 (0.04) & 0.6343 (0.04) & 0.6700 (0.02)\\
Gradient Boosting       & 0.8580 (0.01) & 0.8568 (0.01) & 0.9346 (0.01)\\
Linear SVC              & {\bf 0.8886} (0.01) & {\bf 0.8883} (0.01) & 0.9251 (0.01)\\
Multinomial Naive Bayes & 0.7323 (0.01) & 0.7193 (0.01) & 0.7190 (0.02)\\
Neural Net              & 0.8452 (0.01) & 0.8795 (0.01) & 0.9001 (0.01)\\
Random Forest           & 0.8262 (0.01) & 0.8472 (0.01) & 0.8864 (0.01)\\
\hline
\end{tabular}
\end{table}

\begin{center}
\begin{figure}
   \caption{\label{fig:normalized_cm_multiclass} Normalized Confusion Matrix for multiclass dataset. The normalization is performed per line. A white block indicates that no element has been predicted for the corresponding label. Percentages are reported only if above 1\%.}
   \includegraphics[scale=0.65]{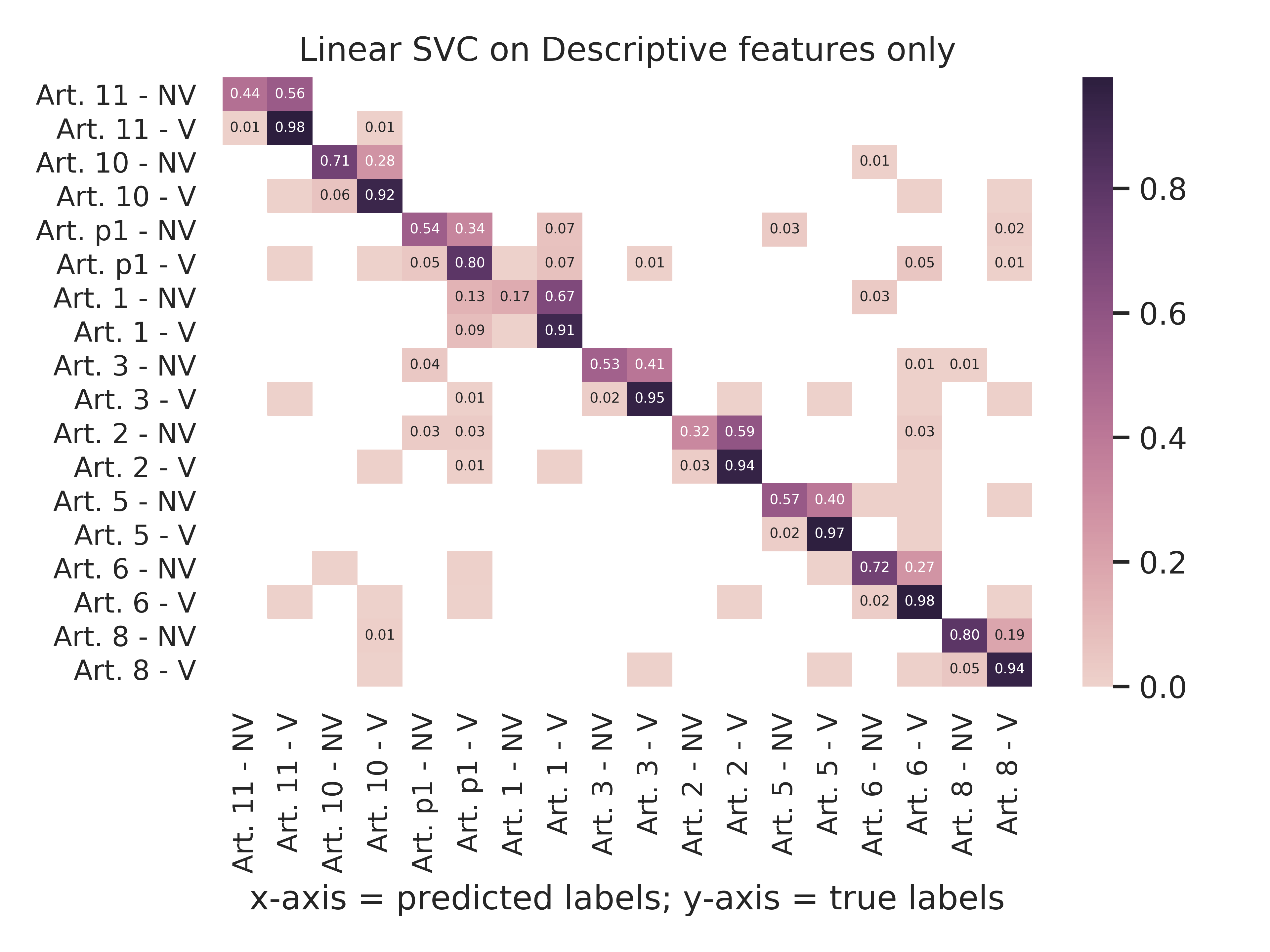}
   \includegraphics[scale=0.65]{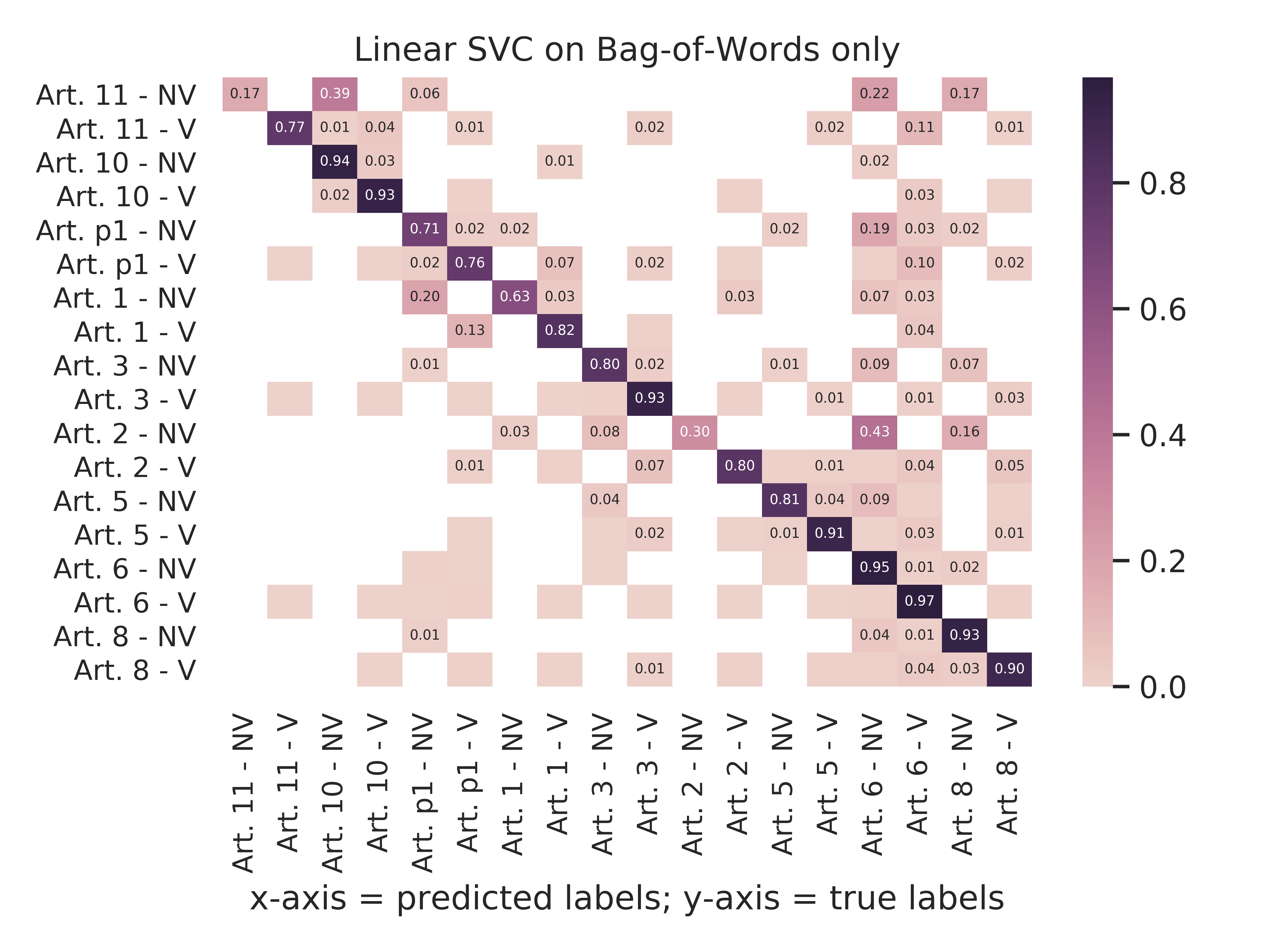}
   \includegraphics[scale=0.65]{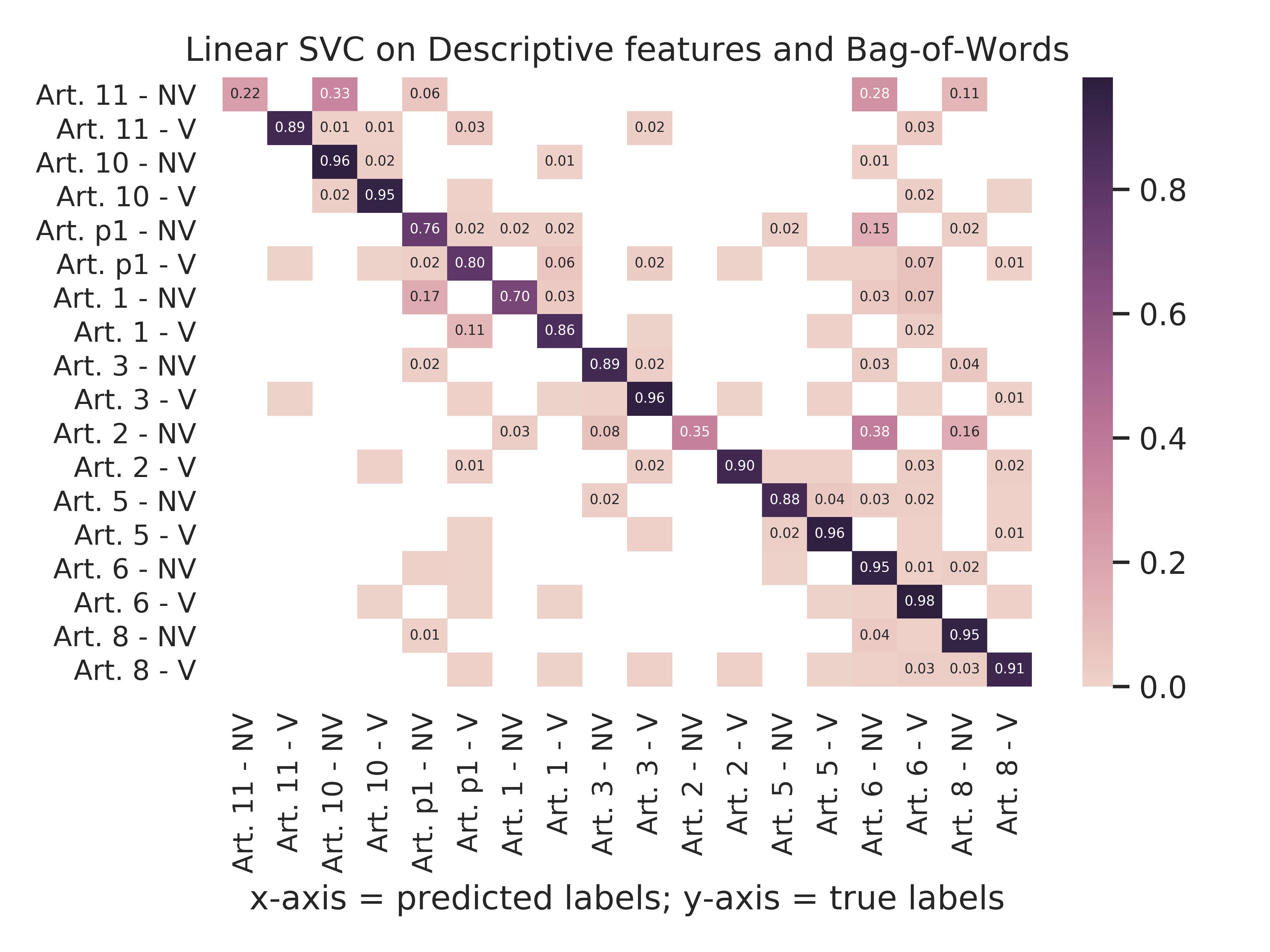}
\end{figure}
\end{center}

Figure \ref{fig:normalized_cm_multiclass} shows the normalized confusion matrix for Linear SVM. The normalization has been done per line, i.e. each line represents the distribution of cases according to their ground truth. For instance, on descriptive features only, for the class "Article 11, no-violation", 44\% only were correctly classified and 56\% assigned to a violation of article 11. The perfect classifier should thus have a diagonal of 1. The diagonal is equivalent to the recall for the corresponding class and the average the diagonal terms is the balanced accuracy (\cite{brodersen2010balanced}).

The flavor ``Descriptive features only'' have a sparser normalized confusion matrix than the counterpart with Bag-of-Words. The fact that the first flavor returns a lower accuracy is explained by looking at the 2x2 blocks on the diagonal. Those blocks are the normalized confusion matrix of the subproblem restricted to find a specific article. For instance, 100\% of non-violation of article 10 has been labeled in one of the two classes related to article 10 (99\% for a violation). In general, the classifiers on ``Descriptive feature only'' are good at identifying the article but generates a lot of false negatives, most likely due to the imbalance between violation and non-violation labels for a given article. Adding the bag-of-words to the case representation slightly lowers the accuracy in average but largely rebalance the 2x2 blocks on the diagonal. On the other hand, it seems that the textual information does not hold enough information to identify the article, which explain why classifiers perform in general lower on this flavor. 

We performed two Wilcoxon signed-rank tests: first between the samples of results on \texttt{BoW} and \texttt{both}, then between \texttt{desc} and \texttt{both}. The first result is clearly significant while the second is not significant, comforting us in the idea that for the multiclass domain, Bag-of-Words alone flavor is unlikely to give good results compared to descriptive features.

The main conclusion to draw from the multiclass experiment is that the descriptive features are excellent at identifying the article while the text offers more elements to predict the judgement. Using only the Bag-of-Words leads to the worst possible results, while adding textual information to the descriptive features slightly increases the accuracy and have a strong beneficial effect on discriminating between violations and non-violations. This is quite in opposition with the conclusion drawn from the experiments on the binary datasets where the textual representation clearly overperformed while the descriptive features had only a marginal effect.

This indicates that it might be more interesting to create a two-stage classifier: a multiclass classifier determines the article based on descriptive features, followed by an article-specific classifier in charge of determining if the article is violated or not. Over-sampling techniques to deal with imbalanced classes constitutes another axis of improvement to explore in future work.

Last but not least, it shows that the benefits of combining sources of information are not monotonic: the best scoring method on individual types of features might not be the best method overall.

\subsection{Multilabel Classification}

The multilabel dataset generalizes the multiclass one in a way there is not only one article to identify before predicting the outcome, but an unknown number. It is closer to real-life situations in which, when a complaint is filled, the precise articles to be discussed are yet to be determined. On top of analyzing the usual performance metrics, we would like to quantify how good are the methods to identify all the articles in each case. From the multiclass results, it is expected that the textual information alone will provide the lowest results among all flavors.

\subsubsection{Protocol}

Appreciating the results of a multilabel classifier is not as easy as in the binary or multiclass case. For instance, having wrongly added one label to 100 cases is not exactly the same as adding 100 wrong labels to a single case. Similarly, being able to predict correctly at least one correct label per case is not the same as predicting all good labels for a fraction of the cases, even if the total amount of correct label is the same in both scenarios. The distributions of ground truth and predicted labels among the dataset are important to evaluate the model.

For this reason, we reported the following multilabel-specific metrics: subset accuracy, precision, recall, F$_1$-score, Hamming loss and the Jaccard Similarity score. The subset accuracy is strictest metric since it measures the percentage of samples such that all the labels are correctly predicted. It does not account for partly correctly labeled vectors. The Jaccard index measures the number of correctly predicted labels divided by the union of prediction and true labels. The Hamming loss calculates the percentage of wrong labels in the total number of labels. Let $T$ (resp. $P$) denotes the true (resp. predicted) labels, $n$ the size of the sample, and $l$ the number of possible labels. Then the metrics are defined by:

\begin{align*}
\text{ACC} & = \frac{1}{n} \underset{i=1}{\overset{n}{\sum}} I(Y_i = \bar Y_i)\\
&\\
\text{RECALL} & = \frac{\text{T} \cap \text{P}}{\text{T}} \\
    & \\
\text{PRECISION} & = \frac{\text{T} \cap \text{P}}{\text{P}} \\
    & \\
\text{F}_1 & = \frac{\text{RECALL} \times \text{PRECISION}}{\text{RECALL} + \text{PRECISION}} \\
    & \\
\text{HAMMING} & = \frac{1}{nl}\underset{i=1}{\overset{n}{\sum}} \text{xor}(y_{i,j}, \bar y_{i, j})
    & \\
\text{JACCARD} & = \frac{\text{T} \cap \text{P}}{\text{T} \cup \text{P}}
\end{align*}

Finally, we are interested in quantifying how much a specific article was properly identified, as well as how much cases with a given number of labels are correctly labeled, taking into account their respective prevalence in the dataset reported by Figure \ref{fig:multilabel_count_labels}. Indeed, about 70\% of cases in the multilabel dataset have only one label such that a classifier assigning only one label to each case could reach about 70\% of subset accuracy.

Not all binary classifiers can be extended for the multilabel problem. We used the five following algorithms: Extra Tree, Decision Tree, Random Forest, Ensemble Extra Tree and Neural Network. As previously, a 10-fold cross-validation has been performed on each flavor.

\subsubsection{Results}

\begin{table}
  \caption{Accuracy obtained for each method on the multilabel dataset.}
  \label{table:multilabel_acc}
  \begin{tabular}{|l|l|l|l| }
\hline
 &  \multicolumn{3}{c|}{Multilabel} \\
\cline{2-4} & desc & BoW & both \\ \hline
Decision Tree       & {\bf 0.7837} (0.02) & 0.6612 (0.02) & {\bf 0.7966} (0.02)\\
Ensemble Extra Tree & 0.7252 (0.03) & 0.6643 (0.02) & 0.6954 (0.02)\\
Extra Tree          & 0.5978 (0.03) & 0.5410 (0.02) & 0.5407 (0.03)\\
Neural Net          & 0.6914 (0.03) & {\bf 0.6745} (0.02) & 0.7159 (0.02)\\
Random Forest       & 0.7061 (0.03) & 0.6438 (0.02) & 0.6674 (0.02)\\
\hline
\end{tabular}
\end{table}

\begin{table}
  \caption{Precision obtained for each method on the multilabel dataset.}
  \label{table:multilabel_precision}
  \begin{tabular}{|l|l|l|l| }
\hline
 &  \multicolumn{3}{c|}{Multilabel} \\
\cline{2-4} & desc & BoW & both \\ \hline
Decision Tree       & 0.8470 & 0.7780 & 0.8705\\
Ensemble Extra Tree & {\bf 0.8808} & {\bf 0.8957} & {\bf 0.9147}\\
Extra Tree          & 0.7202 & 0.6786 & 0.6780\\
Neural Net          & 0.8674 & 0.8719 & 0.8995\\
Random Forest       & 0.8698 & 0.8929 & 0.9120\\
\hline
\end{tabular}
\end{table}

\begin{table}
  \caption{Recall obtained for each method on the multilabel dataset.}
  \label{table:multilabel_recall}
  \begin{tabular}{|l|l|l|l| }
\hline
 &  \multicolumn{3}{c|}{Multilabel} \\
\cline{2-4} & desc & BoW & both \\ \hline
Decision Tree       & {\bf 0.8482} & {\bf 0.7688} & {\bf 0.8611}\\
Ensemble Extra Tree & 0.7635 & 0.7049 & 0.7261\\
Extra Tree          & 0.6999 & 0.6575 & 0.6564\\
Neural Net          & 0.7615 & 0.7671 & 0.7792\\
Random Forest       & 0.7440 & 0.6821 & 0.6996\\
\hline
\end{tabular}
\end{table}

The accuracy is reported in Table \ref{table:multilabel_acc} and shows that Decision Tree outperforms with 79.66\% of cases that have been totally correctly labeled. Similarly to the multiclass setting, the descriptive features provide a better result than the bag-of-words. Decision Tree scores also the best for the F$_1$-score (Table \ref{table:multilabel_f1}) and recall (Table \ref{table:multilabel_recall}). However, Ensemble Extra Tree overperforms Decision Tree when it comes to precision (Table \ref{table:multilabel_precision}). Decision Tree provides the best results for the {\it strict} metrics (highest accuracy and lowest Hamming loss) but also on more permissive metrics (best F$_1$-score and Jaccard index). Therefore, all things being equal (in particular default hyperparameters), Decision Tree is clearly the top method for multilabel which is a bit surprising since it ranked 8th over the binary datasets and 3rd on the multiclass one.

As expected, the bag-of-words flavor provides the worst possible results. Similarly to the experiments on the multiclass dataset, the textual information is inefficient at identifying the article.

\begin{table}
  \caption{$F_1$-score obtained for each method on the multilabel dataset.}
  \label{table:multilabel_f1}
  \begin{tabular}{|l|l|l|l| }
\hline
 &  \multicolumn{3}{c|}{Multilabel} \\
\cline{2-4} & desc & BoW & both \\ \hline
Decision Tree       & {\bf 0.8446} & 0.7711 & {\bf 0.8639}\\
Ensemble Extra Tree & 0.7917 & 0.7720 & 0.7923\\
Extra Tree          & 0.7066 & 0.6651 & 0.6642\\
Neural Net          & 0.7938 & {\bf 0.7991} & 0.8174\\
Random Forest       & 0.7733 & 0.7533 & 0.7720\\
\hline
\end{tabular}
\end{table}

\begin{table}
  \caption{Hamming loss obtained for each method on the multilabel dataset.}
  \label{table:multilabel_hamming}
  \begin{tabular}{|l|l|l|l| }
\hline
 &  \multicolumn{3}{c|}{Multilabel} \\
\cline{2-4} & desc & BoW & both \\ \hline
Decision Tree       & {\bf 0.0188} & 0.0286 & {\bf 0.0169}\\
Ensemble Extra Tree & 0.0195 & 0.0226 & 0.0203\\
Extra Tree          & 0.0350 & 0.0412 & 0.0412\\
Neural Net          & 0.0209 & {\bf 0.0210} & 0.0183\\
Random Forest       & 0.0208 & 0.0239 & 0.0219\\
\hline
\end{tabular}
\end{table}

\begin{table}
  \caption{Jaccard Similarity Score obtained for each method on the multilabel dataset.}
  \label{table:multilabel_jaccard}
  \begin{tabular}{|l|l|l|l| }
\hline
 &  \multicolumn{3}{c|}{Multilabel} \\
\cline{2-4} & desc & BoW & both \\ \hline
Decision Tree       & {\bf 0.8424} & 0.7591 & {\bf 0.8672}\\
Ensemble Extra Tree & 0.7877 & 0.7376 & 0.7652\\
Extra Tree          & 0.6911 & 0.6431 & 0.6452\\
Neural Net          & 0.7673 & {\bf 0.7596} & 0.7888\\
Random Forest       & 0.7700 & 0.7160 & 0.7394\\
\hline
\end{tabular}
\end{table}

\begin{figure}[!h]
\centering
\includegraphics[scale=0.4]{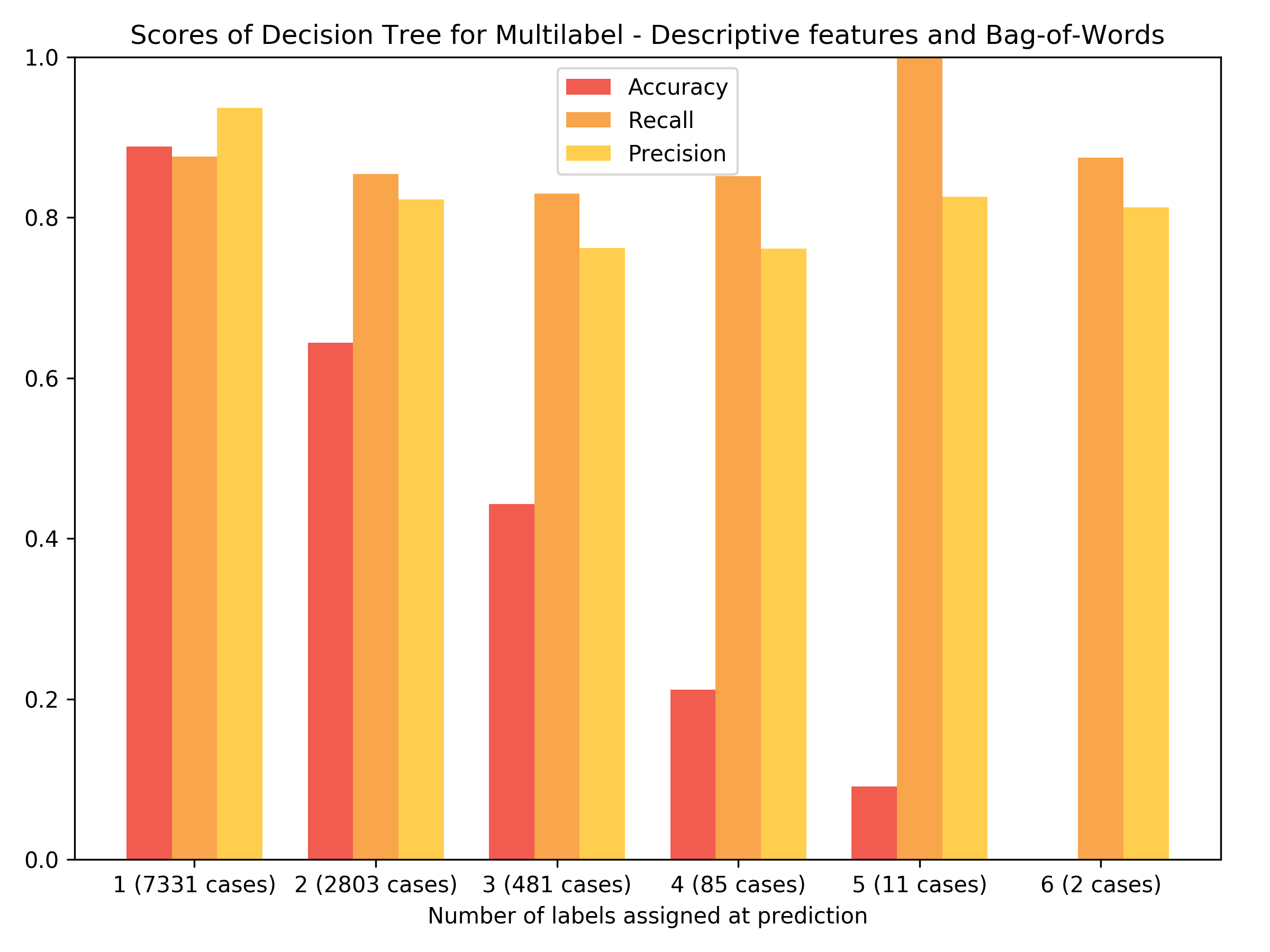}
\caption{{\bf Multilabel scores depending on the number of labels assigned.}}
\label{fig:multilabel_scores}
\end{figure}

The Figure \ref{fig:multilabel_scores} shows the accuracy, recall and precision depending on the number of labels assigned by Decision Tree on the test set. It also indicates the number of cases for each label count. It is striking how the distribution of cases depending on the labels is close to the real distribution described by Figure \ref{fig:multilabel_count_labels}. We can reasonably assume that the model correctly identifies the labels a case is about (or at least the article). The subset accuracy for cases with a single label is consistant with the score on multiclass dataset (which is then virtually similar). The subset accuracy decreases linearly with the number of labels, which is not surprising since the metric become stricter with the number of labels. However, the recall and precision remain stable, above 80\% in average indicating that, not only the algorithm carefully identify the labels (recall) but also identify a large portion of labels (precision). Thus, Figure \ref{fig:multilabel_scores} clearly discards the possibility that the algorithm mostly focuses on cases with a single label.

\section{Conclusion}
\label{sec:conclusion}

In this paper, we introduced the European Court of Human Rights Open Data project consisting in multiple datasets for several variants of the classification problems. The datasets come in different flavors (descriptive features, Bag-of-Word representation, TF-IDF) and are based on the real-life data directly retrieved from the HUDOC database. In total, 13 datasets are provided for the first release. We argued that providing the final data are not enough to ensure quality and trust. In addition, there are always some opinionated choices in the representation, such as the number of tokens, the value of $n$ for the $n$-grams calculation or the weighting schema in the TF-IDF transformation. As a remedy, we provide the whole process of dataset construction from scratch.
The datasets will be iteratively corrected and updated along with the ECHR new judgments. The datasets are carefully versioned to reach a compromise between the need to keep the data up-to-date (as needed by legal practitioners or algorithms in production) and to have the same version of data to compare results between scientific studies.

In the future, we plan to add additional enrichments (e.g. entity extraction from the judgments), new datasets with fine-grain labels and new datasets for different problems (e.g. structured prediction). We hope to offer a web platform such that anyone can tune the different dataset hyperparameters to generate its own flavor: a sort of {\it Dataset as a Service}.

A first experimental campaign has been performed to established a baseline for future work. The predictability power of each dataset and flavor has been tested for the most popular machine learning methods. On binary datasets, we achieved an average accuracy of 0.9443, against 0.9499 for multiclass. It demonstrates the interest of treating the problem at a higher level rather than at the article level. In particular, the learning curves have shown that the models are underfitting on binary datasets but, as the datasets are exhaustive, it is not possible to provide more examples. We showed than the descriptive features are excellent at determining the articles related to a case while the textual features helps in determining the binary outcome. Combining both features always help, but the gap between the two type of features is smaller on the multiclass than on the binary datasets. In both cases, descriptive features actually hold reasonable predictive power. Those preliminary experiments certainly do not clearly answer the realism versus legalism debate, but they open new possibilities to understand better our justice system. They also provide several axes of improvements: hyperparameter tuning, multi-stage classifier and transfer learning.
From those results, it seems clear that the prediction problem can be handled with the current state of the art in artificial intelligence. We hope that this project will pave the way toward a solution to the justification problem.

Last but not least, we encourage all researchers to explore the data, generate new datasets for various problems and submit their contributions to the project. 

\bibliographystyle{spbasic}      


%
%

\section*{Supplementary Material}

\label{S1_Appendix}

\label{S2_Appendix}

\label{S3_Appendix}

\label{S4_Appendix}

\label{S5_Appendix}

\url{https://aquemy.github.io/ECHR-OD_project_supplementary_material/}

\end{document}